\documentclass[fleqn,10pt]{wlscirep}
\usepackage[style=nature,backend=biber]{biblatex}
\bibliography{main.bib}  
\usepackage[utf8]{inputenc}
\usepackage[T1]{fontenc}
\usepackage{amsfonts}
\usepackage{amsmath}
\usepackage{booktabs} 
\usepackage{multirow}
\usepackage{threeparttable}
\usepackage{authblk}
\usepackage{graphicx}  
\usepackage{float}  
\usepackage{tikz}
\usepackage{csquotes}
\usepackage{pifont}
\newcommand{\xmark}{\ding{55}}%

\usepackage{amsmath,amsfonts,bm}

\def\eqref#1{equation~\ref{#1}}

\def\1{\bm{1}}

\def\vx{{\bm{x}}}
\def\vy{{\bm{y}}}
\def\vz{{\bm{z}}}

\DeclareMathAlphabet{\mathsfit}{\encodingdefault}{\sfdefault}{m}{sl}
\SetMathAlphabet{\mathsfit}{bold}{\encodingdefault}{\sfdefault}{bx}{n}

\title{A review of Generative Adversarial Networks for Electronic Health Records: applications, evaluation measures and data sources
}
\author[$\dagger$]{Ghadeer Ghosheh}
\author[$\dagger$]{Jin Li}
\author[$\dagger$]{Tingting Zhu}

\affil[$\dagger$]{Department of Engineering Sciences, University of Oxford}

\affil[ ]{ {\{ghadeer.ghosheh, jin.li, tingting.zhu\}@eng.ox.ac.uk}}

\begin{abstract}
Electronic Health Records (EHRs) are a valuable asset to facilitate clinical research and point of care applications; however, many challenges such as data privacy concerns impede its optimal utilization. Deep generative models, particularly, Generative Adversarial Networks (GANs) show great promise in generating synthetic EHR data by learning underlying data distributions while achieving excellent performance and addressing these challenges. This work aims to review the major developments in various applications of GANs for EHRs and provides an overview of the proposed methodologies. For this purpose, we combine perspectives from healthcare applications and machine learning techniques in terms of source datasets and the fidelity and privacy evaluation of the generated synthetic datasets. We also compile a list of the metrics and datasets used by the reviewed works, which can be utilized as benchmarks for future research in the field. We conclude by discussing challenges in GANs for EHRs development and proposing recommended practices. We hope that this work motivates novel research development directions in the intersection of healthcare and machine learning.

\end{abstract}

\begin{document}





\maketitle

\section{\textbf{Introduction}}
Over the past decade, machine learning (ML) models have proven to have a high potential for supporting medical applications by using data collected in electronic health records (EHRs)~\cite{xiao2018opportunities,sanchez2019machine}. 
Hospitals and medical providers are increasingly adopting and deploying EHR systems. In the US alone, 84\% of hospitals adopted EHR systems as of 2015, which is a 9-fold increase since 2008~\cite{henry2016adoption}. The widespread recording of structured EHRs is paving the way for research opportunities in healthcare applications, such as patient-stratification~\cite{sarraju2021machine}, drug repurposing~\cite{chen2021applications}, public health surveillance~\cite{birkhead2015uses}, as well as the novel discovery of disease mechanisms and correlations as seen in the early COVID-19 applications~\cite{dagliati2021health}. EHRs also provide a valuable asset to develop data-driven and patient-specific clinical decision support systems (CDSS) for diagnostic, prognostic, healthcare cost containment and workflow improvement applications~\cite{sutton2020overview,the2018personalised,shamout2020machine}. However, the full utilization of the wealth of the EHR data in such applications is impeded by several challenges, including data sharing and privacy concerns~\cite{keshta2021security}, where data protection guidelines and regulations such as the
Health Insurance Portability and Accountability Act (HIPAA)~\cite{centers2003hipaa} in the United States, and the General Data Protection Regulation (GDPR)~\cite{voigt2017eu} in Europe have detailed controlling measures that prevent direct access to much of the data for patient privacy purposes. Other data-specific challenges that make EHR processing burdensome include class imbalance~\cite{santiso2019class}, data missingness~\cite{madden2016missing}, noise~\cite{kim2019evolving}, heterogeneity~\cite{conway2011analyzing} and irregular sampling~\cite{sun2020review}. To mitigate these challenges, deep generative models have been proposed to generate synthetic data~\cite{chen2021synthetic}, notably variational autoencoders (VAE)~\cite{kingma2013auto}, and Generative Adversarial Networks (GANs)~\cite{goodfellow2014generative}.

In this paper, we review GANs as one of the most widely-used yet under-studied deep generative frameworks, specifically in the domain of EHR applications. There exist several reviews related to GANs evaluation~\cite{salehi2020generative}, GANs applications for medical imaging~\cite{jeong2022systematic},  and for time-series signals~\cite{zhang2022comprehensive} and observational health data~\cite{georges2020synthetic}. However, in this review, we focus on GANs for structured EHRs, their applications, evaluation and challenges, which serves as a basis for a reading audience with diverse backgrounds. Furthermore, we provide a comprehensive and up-to-date review of the current works and group them based on their target application for healthcare applications, not only for generating synthetic samples, but also mitigating many of the data challenges of EHRs. To the best of our knowledge, this is the first work to discuss and categorise the wide range of evaluation metrics used for evaluating the quality of the synthetic EHRs data generated by GANs. We discuss several open-ended challenges and themes in the current works to motivate new research directions in both the computational and healthcare fields. 
Relevant literature was identified by searching Google Scholar using the keywords "\textit{GAN}" AND "\textit{EHR}", or "\textit{synthetic health data}",  and "\textit{GAN}" AND "\textit{Health}" up until January 2022. We then filtered out papers that used generative models other than GANs, and any duplicate papers.

The outline of the paper is as follows. In section \ref{GAN}, we briefly review the working principles and architecture of GANs and provide an overview of EHR data types in section~\ref{sec:EHR}. We then review the research papers that used GANs for various EHR applications, in section~\ref{Applications}. We discuss and curate a list of commonly used evaluation metrics in section~\ref{eval}, along with the most commonly used data sources in the literature in section~\ref{sources}. We conclude by discussing challenges as well as future directions of GANs for EHRs in section~\ref{next}.

\section{\textbf{GANs: Principles and Architecture}}
\label{GAN}
\begin{figure*}[htpb!]
    \centering
    \resizebox{0.8\textwidth}{!}{
    \includegraphics{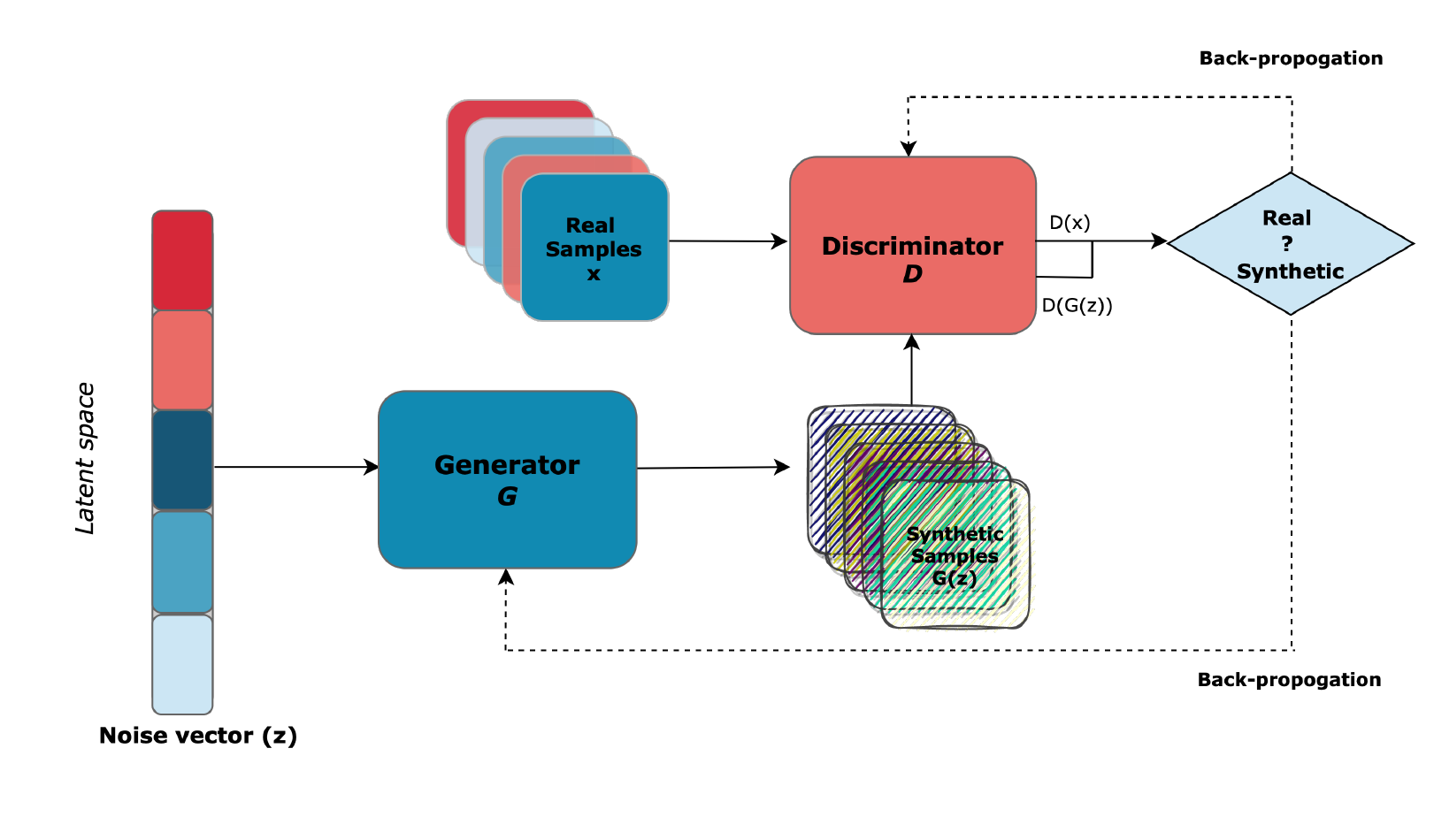}}
    \caption{An overview of the architecture of GANs showing the function of both the \textit{generator} and \textit{discriminator} neural networks. The generator takes an noise vector $\displaystyle \vz$ as input and outputs the synthetic data. The discriminator is trained to distinguish between the real and synthetic data. Both  $\displaystyle G$ and $\displaystyle D$ are then fine-tuned by back-propagation.}
    \label{fig:overview}
\end{figure*}

Since the introduction of GANs in 2014 \cite{goodfellow2014generative}, they have shown great potential in generating realistic data for various applications. The working principle of GANs essentially involves the training of a pair of deep neural networks in competition with each other~\cite{creswell2018generative}. The first neural network, the \textit{generator} $\displaystyle G$, takes a noise vector $\displaystyle \vz$ from latent space as an input and generates the synthetic samples $\displaystyle G(\vz)$~\cite{creswell2018generative}. While the other neural network, the \textit{discriminator}, $\displaystyle D$ is given both the real $\displaystyle \vx$ and generated samples $\displaystyle G(\vz)$, and is trained to discriminate between the real and synthetic ones~\cite{goodfellow2014generative}. The discriminator outputs a vector of probability predictions of whether the inputted samples were real or synthetic. Both the generator and discriminator are fine-tuned using the discriminator's output via back-propagation as shown in Figure~\ref{fig:overview}. The training involves both finding the parameters of a discriminator that maximize its classification accuracy and finding the parameters of a generator that minimize the discriminator's ability to tell the real and synthetic samples apart~~\cite{goodfellow2014generative}. In other words, the objective loss function of GANs is:
\begin{align*}
\min_{G}\max_{D} V(D,G) =\ \mathbb{E}_{\mathrm{\displaystyle \vx}}[\log D(\mathrm{\displaystyle \vx})] + \mathbb{E}_{\mathrm{\displaystyle \vz}}\left[\log({1-D(G(\displaystyle \vz))}\right)]
\end{align*}

The initial results of GANs were promising~\cite{goodfellow2014generative}, which motivated researchers to propose modifications and adaptations for specific tasks and applications. Notably,~\cite{mirza2014conditional}, proposed the Conditional Generative Adversarial Nets (\textit{CGAN}), which generated data by conditioning the GAN on a selected variable or label $\displaystyle \vy$ , where $\displaystyle \vy$ is fed to the generator and discriminator as input layer. Another important work is deep convolutional GAN (\textit{DCGAN})  which utilized a pair of deep convolutional networks for each of the \textit{G} and \textit{D}~\cite{radford2015unsupervised}. Around the same time, an Information Maximizing Generative Adversarial Networks (\textit{InfoGAN}) was proposed to provide additional interpretability where semantic meaning was introduced to the variables in the latent space~\cite{chen2016infogan}. Recurrent GAN (\textit{RCGAN})~\cite{esteban2017real}, extended the original GAN model to generate sequential data by using recurrent neural networks (RNN) for EHR applications,  motivating several GAN applications for time-series data. Wasserstein Generative Adversarial Networks (\textit{WGAN}) was introduced with the main contribution in modifying the loss function to improve GAN training stability by using the Wasserstein distance metric to measure the distribution similarity of real and synthetic data~\cite{arjovsky2017wasserstein1,gulrajani2017improved}. Other important works include \textit{(CycleGAN)}~\cite{zhu2017unpaired} and \textit{(STARGAN)}~\cite{choi2018stargan}, which were adapted to allow for domain translation, and diversity sensitive conditional GAN \textit{(DSCGAN)} which regularizes the generator to produce diverse outputs~\cite{yang2019diversity}.

Despite their high potential, training GANs involves many challenges, notably mode collapse. Mode collapse refers to the case where the generator maps different inputs to a small set of synthetic outputs, rather than producing diverse outputs that reflect the input~\cite{goodfellow2014generative}. Another challenge is vanishing gradients~\cite{arjovsky2017towards}, where the discriminator is performing very well and not providing useful information to improve the generator training, leading the generator's gradient to vanish~\cite{arjovsky2017towards}. To address these challenges, some architectures and modifications to the loss function were proposed as seen in \textit{WGAN}~\cite{arjovsky2017wasserstein1,gulrajani2017improved}, minibatch discrimination~\cite{salimans2016improved, goodfellow2016nips}, minibatch averaging~\cite{choi2016multi}, unrolled GANs~\cite{metz2016unrolled}, and noise injection~\cite{salimans2016improved}. Notwithstanding the advantages of GANs, improving GAN training stability remains one of the bottlenecks in scaling GAN applications in real-world settings. 


\section{\textbf{EHRs: Data Types and Clinical Settings}}
\label{sec:EHR}
\begin{figure*}[htpb!]
    \centering
    \resizebox{0.85\textwidth}{!}{
    \includegraphics{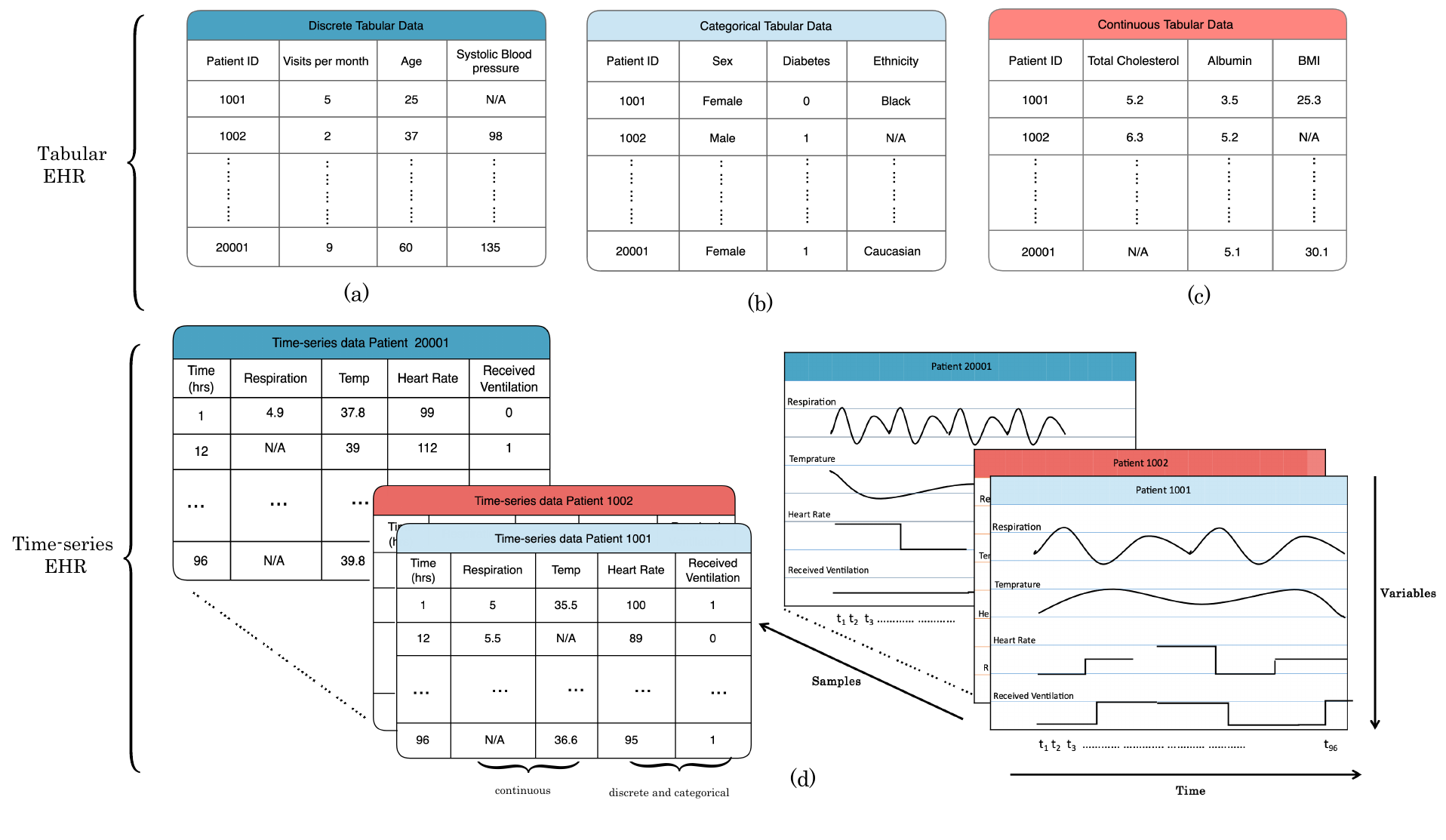}}
    \caption{The two main types of EHR data, tabular and time-series are shown in their various forms. Discrete, categorical and continuous tabular data are shown in (a), (b) and (c), respectively. Time series data is shown in (d), where the record is shown on the left and a corresponding plot of the data is shown on the right.}
    \label{fig:EHRtypes}
\end{figure*}

In medical practice, medical staff use EHRs to record and capture various forms of data about a patient during an encounter. Like paper records, EHRs store data such as hospitalization information, and patient-level information such as demographics, comorbidities, medical history, vital signs, laboratory tests,  prescribed medication, administered interventions, diagnosis, and clinical outcomes~\cite{birkhead2015uses}. The nature of each of these kinds of data differs, which results in multiple types of EHR data. Structured EHR data can be presented in either tabular or time-series formats, as shown in Figure~\ref{fig:EHRtypes}. Tabular data stores information that presents a representation of the patient's encounter such as demographic features, aggregated mean or a one time measurement of vital signs,  where each sample has one value for each feature.  Time-series data, on the other hand, presents a record of data points indexed in time order, which might be used to present disease progression over time as seen in longitudinal data~\cite{herrett2015data} or even short-term records as seen in vital signs~\cite{sornmo2006electrocardiogram}.  The variables recorded in each of the two data types can be either discrete, categorical, or continuous. Discrete variables represent values that can be obtained by counting and stored as integers such as age or number of visits per month, as seen in Figure~\ref{fig:EHRtypes} (a) and Figure~\ref{fig:EHRtypes} (d). Categorical variables, on the other hand, are used when there is a finite number of categories such as sex or ethnicity, as shown in as seen in Figure~\ref{fig:EHRtypes} (b) and Figure~\ref{fig:EHRtypes} (d). Lastly, continuous variables are variables whose value is obtained by measurement and is not limited to whole numbers. Examples of continuous variables can be seen in many laboratory tests and vital signs such as albumin, body temperature, and total cholesterol, as shown in Figure~\ref{fig:EHRtypes} (c) and Figure~\ref{fig:EHRtypes} (d). It is worth noting, that different EHR data types usually coexist in the same patient record. For example, a patient might have both tabular and time-series data recorded for the same visit. This heterogeneous nature of EHRs often results in complexity in terms of its analysis, modeling, and use for machine learning purposes~\cite{conway2011analyzing,xiao2018opportunities}. 

EHR data can be recorded in different settings and stages of a patient encounter or observation.  During a hospital visit, a patient encounter can be classified as either inpatient or outpatient, where the first requires hospitalization and admission, while the latter does not. For an inpatient encounter, a patient could go through various units within the same facility, which depends on the clinical status~\cite{young2003inpatient}, availability of human and material resources~\cite{de2018patient}, or hospital capacity~\cite{fekieta2021organisational}.  At the beginning of a hospital presentation, patients can be presented to the emergency wards where initial diagnosis and interventions take place~\cite{teich1998information}, where the focus is to admit and then triage the patient based on the medical need. In the general inpatient-wards, patients get regular laboratory tests, vital sign checks, treatment administration, and other required procedures as requested by the doctor. Patients who deteriorate or those whose cases require higher care are admitted to the Intensive Care Unit (ICU), where the data tends to be frequently collected as the patient is under close monitoring. Data collected in ICUs, are usually referred to as critical care data~\cite{imhoff1992acquisition}. The other type of EHR data is that of outpatient encounters, where the data collected is for patients who were not admitted to the hospital, as seen in the case of specialist consultations~\cite{helmer2008primary} and visits to general practitioners~\cite{crpd}. The nature of outpatient data varies across countries, depending on the availability of primary care and the need for referrals to get a specialist consultation.


\section{\textbf{Application of GANs for EHRs}}
\label{Applications}
The applications of GANs in the medical domain are very diverse, specifically in medical imaging. For instance, GANs have been used for various radiology tasks that ranged from data augmentation to data segmentation and denoising~\cite{malygina2019data,zhang2020unsupervised,yi2018sharpness}. However, there is much less work on using GANs to generate realistic structured healthcare data such as EHRs. The lag in the use of GANs for EHR data can be attributed to the many data challenges, such as complexity, heterogeneity,  missingness~\cite{xiao2018opportunities}. In comparison with other data modalities such as images and text, which can be intuitively and visually evaluated for realism, assessing the quality of the generated EHR data is difficult. In Table \ref{tab:EHR_GAN}, we summarise major works that used GANs for EHR applications and group them based on their target application. The main groups are (1) generation of diverse types of EHRs, (2) semi-supervised learning and data augmentation, (3) imputation of missingness, (4) treatment effect estimation, and (5)  privacy-preservation. The works are reviewed in terms of the used models, task, dataset size, open access code and data as well as evaluation components used to assess the quality of the synthetic data. 
 \begin{table*}[htpb!]
 \caption{Summary of the various uses of GANs for EHRs and comparison of target application, evaluation measures, medical datasets and open access.}
 \resizebox{\textwidth}{!}{
  \begin{threeparttable}[b]

  \centering

 \begin{tabular}{llllllllllllllll}
\toprule
& & \multicolumn{2}{c}{\textbf{Problem}} & \multicolumn{7}{c}{\textbf{Evaluation}} & \multicolumn{2}{c}{\textbf{Medical Dataset}} & \multicolumn{2}{c}{\textbf{Open Access}}\\

\cmidrule{3-4}
\cmidrule(lr{1em}){5-11}
\cmidrule(lr{1em}){12-13}
\cmidrule{14-15}
 \multirow{1}*{References} & \multirow{1}*{Year} & \multirow{1}*{Model} & \multirow{1}*{Task}  & DWS & LDS & JDS & IDRS & PP & DU & Qual &  \multirow{1}*{Dataset} & \multirow{1}*{Dataset size (N/R)} & \multirow{1}*{Dataset} & \multirow{1}*{Code}\\

 \midrule
\textbf{Generation of Diverse types of EHRs} & &&&&&&&&&&&&\\
\midrule
\multirow{3}*{\cite{choi2017generating}} & \multirow{3}* {2017} & \multirow{3}* {medGAN} & \multirow{3}*{Generating discrete tabular EHR data} &  \multirow{3}*{\checkmark} & \multirow{3}*{\xmark} & \multirow{3}*{\xmark} & \multirow{3}*{\checkmark} &\multirow{3}*{\checkmark} & \multirow{3}*{\xmark} &\multirow{3}*{\checkmark} & MIMIC-III & 46,520/ NA & \checkmark*  & \multirow{3}*{\checkmark}\\ 
        & &&&&&&&&& & Sutter PAMF & 258,559/ NA & \xmark &\\ 
         & &&&&&&&&&& Sutter Heart failure Cohort  & 30,738/ NA & \xmark &\\ 

\midrule
\cite{esteban2017real} & 2017 & RGAN, RCGAN &  Generating continuous time-series EHRs & \xmark  & \xmark & \checkmark & \xmark & \checkmark & \checkmark & \checkmark & Philips eICU  & 17,693/ NA & \checkmark* & \checkmark\\
\midrule
\cite{yahi2017generative} & 2017 & GAN for DLEs & Generating  continuous time-series Drug Laboratory Effects (DLEs) & \xmark  & \xmark & \checkmark & \xmark &  \xmark &  \xmark & \checkmark & Private New York EHR dataset & 4,830/ NA & \xmark & \xmark\\
 \midrule
 
\cite{yoon2018radialgan} & 2018 & RadialGAN & Leveraging multiple tabular datasets by using multiple GAN & \xmark & \xmark & \xmark & \xmark & \xmark & \checkmark & \xmark& 14 RCTs from MAGGIC & 528-13279/ NA & \checkmark* & \checkmark \\
\midrule
\multirow{2}*{\cite{baowaly2019synthesizing}} &\multirow{2}*{2019}  & \multirow{2}*{medWGAN, medBGAN} & \multirow{1}*{Itegrating medGAN with WGAN-GP \& BGAN  for generating }  & \multirow{2}*{\checkmark} &\multirow{2}*{\xmark}  & \multirow{2}*{\xmark} & \multirow{2}*{\checkmark} & \multirow{2}*{\xmark} & \multirow{2}*{\xmark} & \multirow{2}*{\xmark} & MIMIC-III & 46,517 / 46,517 & \checkmark* &\multirow{2}*{\checkmark}\\ 
        && & discrete tabular EHRs  &&&&&& && NHIRD Taiwan & 498,909 / 498,909 & \xmark &\\ 
\midrule
\multirow{1}* {\cite{chin2019generation}} & \multirow{1}*{2019} & \multirow{1}*{WGAN} &\multirow{1}*{Generating heterogeneous discrete tabular EHRs} &  \multirow{1}*{\checkmark} &  \multirow{1}*{\xmark}  & \multirow{1}*{\checkmark} &  \multirow{1}*{\xmark}  & \multirow{1}*{\checkmark} & \multirow{1}*{\checkmark} &  \multirow{1}*{\checkmark} & NMDS & NA / 2,873,466 & \xmark & \multirow{1}*{\xmark}  \\
\midrule

 \multirow{2}*{\cite{Scgan}}&\multirow{2}* {2019}  &\multirow{2}* {SC-GAN} &  Generating continuous sequentially coupled time-series EHRs & \multirow{2}*{\checkmark} & \multirow{2}*{\xmark} & \multirow{2}*{\xmark} & \multirow{2}*{\checkmark} & \multirow{2}*{\xmark} & \multirow{2}*{\checkmark} & \multirow{2}*{\checkmark} & \multirow{2}* {MIMIC-III} & \multirow{2}* {29,278 / NA} & \multirow{2}* {\checkmark*} & \multirow{2}*{\xmark} \\ 
  & &&  data for patient sate \& medication dosage & &  &  &  & & & &&& &\\
  \midrule
  \multirow{1}*{\cite{zhang2020ensuring}} &  \multirow{1}*{2020} & \multirow{1}*{EMR-WGAN, EMR-CWGAN} &  Improved EHR generation training stability and evaluation &  \multirow{1}*{\checkmark}  & \multirow{1}*{\checkmark}  & \multirow{1}*{\checkmark} & \multirow{1}*{\checkmark}  & \multirow{1}*{\checkmark}  & \multirow{1}*{\xmark}& \xmark & \multirow{1}*{VUMC Synthetic Derivative} & \multirow{1}*{2,246,444 / NA} & \multirow{1}*{\xmark} & \multirow{1}*{\xmark} \\
  \midrule
  \multirow{1}*{\cite{goncalves2020generation}} & \multirow{1}*{2020} & \multirow{1}*{MC-medGAN} & Generating multi-categorical tabular EHRs &  \multirow{1}*{\checkmark}  &  \multirow{1}*{\checkmark}  &  \multirow{1}*{\xmark} &  \multirow{1}*{\checkmark} &  \multirow{1}*{\checkmark} & \multirow{1}*{\xmark} & \checkmark & \multirow{1}*{SEER’s research dataset} & \multirow{1}*{NA / 366,631} & \multirow{1}*{\checkmark*} & \multirow{1}*{\checkmark} \\
\midrule

 \multirow{2}*{\cite{yan2020generating}} &  \multirow{2}*{2020}  & \multirow{2}*{HGAN} & Generating Heterogeneous tabular EHRs while preserving feature & \multirow{2}*{\checkmark}  & \multirow{2}*{\checkmark}  & \multirow{2}*{\checkmark}  & \multirow{2}*{\checkmark}  & \multirow{2}*{\checkmark}  & \multirow{2}*{\xmark} & \multirow{2}*{\checkmark} &   \multirow{2}*{VUMC Synthetic Derivative} & \multirow{2}*{928,089 / NA} &  \multirow{2}*{\xmark} &  \multirow{2}*{\xmark}\\
  &&& constraints and inter-dimensional dependencies &&& &&&&&&&&\\
\midrule

 \multirow{2}*{\cite{torfi2020cor}} & \multirow{2}*{2020} &  \multirow{2}*{CorGAN} &  \multirow{1}*{Correlation-Capturing generation of continuous and discrete} & \multirow{2}*{\checkmark} & \multirow{2}*{\xmark}   &  \multirow{2}*{\xmark} & \multirow{2}*{\checkmark} & \multirow{2}*{\checkmark} & \multirow{2}*{\checkmark}  &  \multirow{2}*{\xmark} & MIMIC-III & NA / 46,000 & \checkmark* & \multirow{2}*{\checkmark}\\
  && & tabular EHRs & & & &&&&& UCI Epileptic Seizure Recognition & 500 / 11,500 & \checkmark\\
 \midrule
 
\cite{zhang2021synteg} & 2021 & SynTEG &  Generating discrete time-series EHRs of diagnostic events &  \checkmark & \checkmark &  \xmark  & \xmark &\checkmark & \checkmark & \xmark& VUMC Synthetic Derivative & 
NA / 2,187,629 & \xmark &\checkmark \\
\midrule 

 \multirow{2}*{\cite{lee2020generating}} & \multirow{2}*{2020} &  \multirow{2}*{DualAAE} &  \multirow{2}*{Generating time-series EHRs of discrete diagnostic codes } & \multirow{2}*{\xmark} & \multirow{2}*{\xmark} &  \multirow{2}*{\checkmark} & \multirow{2}*{\xmark} & \multirow{2}*{\checkmark} & \multirow{2}*{\xmark} &  \multirow{2}*{\checkmark} & 
 MIMIC-III & 7,537 / 19,993  & \checkmark* & \multirow{2}*{\checkmark}\\
  && & & & & &&&&&
  UT-Physicians & 13,025 / 85,845 & \xmark\\
 \midrule

\multirow{3}*{\cite{li2021generating}} & \multirow{3}* {2021} & \multirow{3}* {EHR-M-GAN} & \multirow{3}*{Generating mixed-type time-series EHRs} &  \multirow{3}*{\checkmark} & \multirow{3}*{\xmark} & \multirow{3}*{\checkmark} & \multirow{3}*{\checkmark} &\multirow{3}*{\checkmark} & \multirow{3}*{\checkmark} &\multirow{3}*{\checkmark} & MIMIC-III & 28,344 /  28,344 & \checkmark* & \multirow{3}*{\checkmark}\\ 
& &&  &&&&&&& & 
Philips eICU & 99,015 / 99,015 & \checkmark* &\\
& &&  &&&&&&& & 
HiRID & 14,129 / 14,129 & \checkmark* &\\ 
\midrule


\multirow{1}*{\cite{weldon2021generation}} & \multirow{1}*{2021} &  \multirow{1}*{WGAN} & \multirow{1}*{Federated learning for GAN for discrete, binary EHRs} & \multirow{1}*{\checkmark} & \multirow{1}*{\xmark}   &  \multirow{1}*{\checkmark} & \multirow{1}*{\xmark} & \multirow{1}*{\checkmark} & \multirow{1}*{\xmark}  & \multirow{1}*{\checkmark}& \multirow{1}*{MIMIC-III} & \multirow{1}*{NA / 46,520} & \multirow{1}*{\checkmark*}  & \multirow{1}*{\xmark} \\
\midrule

\textbf{Semi-Supervised Learning and Data Augmentation}& &&&&&&&&&&&&& \\
\midrule
\multirow{2}*{\cite{che2017boosting}} & \multirow{2}*{2017} & \multirow{2}*{ehrGAN,SSL-GANs} & Augmenting data for imbalanced SSL tasks using EHRs & \multirow{2}*{\xmark} &\multirow{2}*{\xmark}  & \multirow{2}*{\xmark} & \multirow{2}*{\checkmark} &  \multirow{2}*{\xmark} & \multirow{2}*{\checkmark} &\multirow{2}*{\checkmark}& \multirow{2}*{Private insurance dataset} &  \multirow{2}*{218,680 / 14,969,489} &\multirow{2}*{\xmark} & \multirow{2}*{\xmark}\\
  & && of sequences of diagnosis codes &&&&&&&& & & &\\ 
\midrule
\multirow{2}*{\cite{li2018semi}} & \multirow{2}*{2018} & \multirow{2}*{GAN for SSL} & SSL based GANs for detecting rare diseases in unlabelled& \multirow{2}*{\xmark}& \multirow{2}*{\xmark} & \multirow{2}*{\xmark} & \multirow{2}*{\xmark} & \multirow{2}*{\xmark} & \multirow{2}*{\checkmark}& \multirow{2}*{\xmark} & \multirow{2}*{IQVIA Rx \& Dx} & \multirow{2}*{2,961,750 / NA} & \multirow{2}*{\xmark} & \multirow{2}*{\checkmark} \\
  & && tabular discrete \& continuous EHRs  &&&&&&&& & &  &\\ 
\midrule
\multirow{2}*{\cite{yu2019rare}} & \multirow{2}*{2019} & \multirow{2}*{GAN for SSL} & SSL based GANs for detecting rare diseases in unlabelled& \multirow{2}*{\xmark}& \multirow{2}*{\xmark} & \multirow{2}*{\xmark} & \multirow{2}*{\xmark} & \multirow{2}*{\xmark} & \multirow{2}*{\checkmark}  & \multirow{2}*{\xmark} & \multirow{2}*{IQVIA Rx \& Dx} & \multirow{2}*{1,792,760 / NA} & \multirow{2}*{\xmark} & \multirow{2}*{\xmark} \\
  & && time-series EHRs &&&&&&& & & & &\\ 
\midrule
\multirow{2}*{\cite{yang2019gan}} & \multirow{2}*{2019} & \multirow{2}*{GAN} & \multirow{1}*{SSL based labeling of unlabled data, and GAN-based } &  \multirow{2}*{\xmark} & \multirow{2}*{\xmark} &  \multirow{2}*{\xmark} & \multirow{2}*{\xmark} & \multirow{2}*{\xmark} &  \multirow{2}*{\checkmark} &  \multirow{2}*{\checkmark} & 20 datasets from UCI  &  NA / 80-2,000 & \checkmark & \multirow{2}*{\xmark}\\ 
      & & & data augmentation in tabular EHRs & & & & & && & Cerebral stroke private dataset & 11,039 / NA & \xmark & \\ 
\midrule
\textbf{Imputation of Missingness}& &&&&&&&&&&&&&\\
\midrule
\cite{yoon2018gain} & 2018 & GAIN & GAN-based discrete \& categorical tabular data imputation & \xmark & \xmark & \checkmark & \xmark & \xmark & \checkmark & \xmark  & UCI Breast dataset  & NA / 569 & \checkmark & \checkmark\\ 
\midrule
\multirow{2}*{\cite{zhang2018medical}} &  \multirow{2}*{2018} &  \multirow{2}*{Stackelberg GAN} & \multirow{1}*{Stabilizing GAIN imputation for discrete, continuous, \&} & \multirow{2}*{\xmark} & \multirow{2}*{\xmark} & \multirow{2}*{\xmark} & \multirow{2}*{\xmark} & \multirow{2}*{\xmark} & \multirow{2}*{\checkmark} & \xmark & \multirow{2}*{MIMIC-III} & \multirow{2}*{38,645 / 58,000} & \multirow{2}*{\checkmark*} & \multirow{2}*{\xmark}\\ 
      & & &   categorical EHRs using Stackelberg principles & & & & & & & & & & &\\ 
\midrule
\cite{luo2018multivariate} & 2018 & GAN with GRUI & GAN-based multivariate time-series EHR imputation & \xmark & \xmark & \xmark & \xmark & \xmark & \checkmark & \xmark  & PhysioNet Challenge 2012 & NA/ 4,000 & \checkmark & \checkmark\\ 
\midrule
\cite{luo2019e2gan} & 2019 & E\textsuperscript{2}GAN & Improved GAN-based multivariate time-series EHR imputation & \xmark & \xmark & \xmark & \xmark & \xmark & \checkmark & \xmark  & PhysioNet Challenge 2012 dataset& NA/ 4,000 & \checkmark & \checkmark\\ 
\midrule
\multirow{2}*{\cite{yang2019categorical}} &\multirow{2}*{2019} &  \multirow{2}*{Categorical GAIN} & \multirow{2}*{Improving GAIN imputation of categorical tabular EHRs} &\multirow{2}*{\xmark} & \multirow{2}*{\xmark} & \multirow{2}*{\xmark} & \multirow{2}*{\xmark} & \multirow{2}*{\xmark} &  \multirow{2}*{\checkmark} &\multirow{2}*{\xmark} & UCI Breast Cancer &  NA/ 286 & \checkmark &  \multirow{2}*{\xmark} \\ 
 && && & & & & & & & PRAEGNANT study & 1234 / NA  & \xmark & \\
 
 \midrule
\multirow{2}*{\cite{camino2019improving}} &\multirow{2}*{2019} &  \multirow{1}*{GAIN adaptation} & \multirow{1}*{Improving GAIN imputation of mixed tabular EHRs, including } &\multirow{2}*{\xmark} & \multirow{2}*{\xmark} & \multirow{2}*{\xmark} & \multirow{2}*{\xmark} & \multirow{2}*{\xmark} &  \multirow{2}*{\checkmark} & \multirow{2}*{\xmark} & \multirow{2}*{UCI breast dataset}  & \multirow{2}*{NA / 569} & \multirow{2}*{\checkmark} &  \multirow{2}*{\checkmark} \\ 

    & & &  multi-categorical features & & & && && & & &\\
 \midrule

  \multirow{2}*{\cite{dai2021multiple}} & \multirow{2}*{2021} &  \multirow{2}*{MI-GAN}  &   \multirow{1}*{GAN-based multiple imputation for} &  \multirow{2}*{\xmark} & \multirow{2}*{\xmark}  & \multirow{2}*{\xmark} & \multirow{2}*{\xmark} & \multirow{2}*{\xmark} &   \multirow{2}*{\checkmark} & \multirow{2}*{\xmark}  & \multirow{2}*{ADNI dataset} & \multirow{2}*{ NA / 649} & \multirow{2}*{\checkmark*} & \multirow{2}*{\xmark}\\ 
      & & &  categorical time-series EHRs & & & && && & &&  \\
 \midrule
 \multirow{2}*{\cite{gupta2021concurrent}} & \multirow{2}*{2021} &  \multirow{2}*{Bi-GAN}  &   \multirow{2}*{Concurrent imputation and prediction in time-series EHRs} &  \multirow{2}*{\xmark} & \multirow{2}*{\xmark}  & \multirow{2}*{\xmark} & \multirow{2}*{\xmark} & \multirow{2}*{\xmark} &   \multirow{2}*{\checkmark} & \multirow{2}*{\xmark}  & Nemours Pediatric  & 66,878 / NA & \checkmark* & \multirow{2}*{\checkmark}\\ 
      & & &  & & & && && & All of Us & 34,226 / NA & \checkmark \\
 \midrule

\textbf{Treatment Effect Estimation} & &&&&&&&&&&&&&\\
  \midrule
  
\multirow{2}*{\cite{yoon2018ganite}}& \multirow{2}*{2018} & \multirow{2}*{GANITE} &  \multirow{1}*{Generating missing counterfactual data and individualized}  & \multirow{2}*{\xmark}  & \multirow{2}*{\xmark} & \multirow{2}*{\xmark}  & \multirow{2}*{\xmark} & \multirow{2}*{\xmark} & \multirow{2}*{\checkmark} & \multirow{2}*{\xmark} & Twins & 11,400 / 11,400  & \checkmark & \multirow{2}*{\checkmark}\\
 & &&treatment effects estimation in tabular EHRs & & &  &  &  &  && IHDP  & 747 / 747  & \checkmark &\\
  \midrule

  \multirow{2}*{\cite{mcdermott2018semi}} & \multirow{2}*{2018} & \multirow{2}*{CWR-GAN}& Generating time-series post-treatment outcomes for ITE & \multirow{2}*{\xmark} & \multirow{2}*{\xmark} & \multirow{2}*{\xmark} & \multirow{2}*{\xmark} & \multirow{2}*{\xmark} & \multirow{2}*{\checkmark} & \multirow{2}*{\xmark} & \multirow{2}*{MIMIC-III} & \multirow{2}*{2,000 / NA } & \multirow{2}*{\checkmark*} & \multirow{2}*{\checkmark} \\
     & && estimation in biomedical translation tasks &&&&&&&&&&&\\

\midrule
\multirow{2}*{\cite{ge2020conditional}}& \multirow{2}*{2020} & \multirow{2}*{MGANITE} &  Estimating effects of continuous, binary and categorical &  \multirow{2}*{\xmark}&   \multirow{2}*{\xmark}  &  \multirow{2}*{\xmark}&   \multirow{2}*{\xmark}   &  \multirow{2}*{\xmark} &  \multirow{2}*{\checkmark} &   \multirow{2}*{\xmark} &\multirow{2}*{AML dataset}  &  \multirow{2}*{NA/212} & \multirow{2}*{\checkmark*} & \multirow{2}*{\checkmark}\\
& && treatments via conditional GANs on tabular EHRs & & & & &&  & & &  & & \\
  \midrule

  \multirow{2}*{\cite{li2020continuous}} &   \multirow{2}*{2020} &   \multirow{2}*{GAD} & Continuous treatment effect estimation by deconfounding &  \multirow{2}*{\xmark}&   \multirow{2}*{\xmark} & \multirow{2}*{\xmark} &  \multirow{2}*{\xmark} &  \multirow{2}*{\xmark} & \multirow{2}*{\checkmark} & \multirow{2}*{\xmark}  &  \multirow{2}*{Twins} &   \multirow{2}*{4,821 / NA} &   \multirow{2}*{\checkmark} &   \multirow{2}*{\xmark}  \\
        & && in tabular EHRs &&&&&&&&&&&\\
\midrule
\multirow{2}*{\cite{ghosh2021propensity}}& \multirow{2}*{2021} & \multirow{2}*{PSSAM-GAN} &  \multirow{2}*{Propensity score augmentation matching for tabular EHRs} &  \multirow{2}*{\xmark}&   \multirow{2}*{\xmark}  &  \multirow{2}*{\xmark}&   \multirow{2}*{\xmark}   &  \multirow{2}*{\xmark} &  \multirow{2}*{\checkmark} &  \multirow{2}*{\xmark} & S. aureus dataset  &  NA / 2,006 & \xmark & \multirow{2}*{\checkmark}\\
& &&  & & &  &&  & && IHDP  & 747 / 747 & \checkmark & \\
\midrule
 \textbf{Privacy Preservation} & &&&&&&&&&&&&&\\
 \midrule
  \multirow{2}*{\cite{xie2018differentially}} & \multirow{2}*{2018} & \multirow{2}*{DPGAN}& Generating differential private EHR data using moment-& \multirow{2}*{\checkmark} & \multirow{2}*{\xmark} & \multirow{2}*{\xmark} & \multirow{2}*{\checkmark} & \multirow{2}*{\checkmark} & \multirow{2}*{\xmark} & \multirow{2}*{\xmark} & \multirow{2}*{MIMIC-III} & \multirow{2}*{NA / 46,520} & \multirow{2}*{\checkmark*} & \multirow{2}*{\checkmark} \\
     & && accounting techniques &&&&&&&&&&&\\
\midrule
 \multirow{4}*{\cite{jordon2018pate}} & \multirow{4}*{2018} &  \multirow{4}*{PATE-GAN} & \multirow{4}*{Generating differential private tabular data using PATE}  &\multirow{4}*{\xmark}  & \multirow{4}*{\xmark} & \multirow{4}*{\xmark}  & \multirow{4}*{\xmark}  & \multirow{4}*{\checkmark}  & \multirow{4}*{\checkmark}  & \multirow{4}*{\xmark} & UCI Epileptic Seizure Recognition  & NA / 11,500 & \checkmark& \multirow{4}*{\checkmark}\\
    &&& &  & & & &  & & & Kaggle Cervical Cancer & NA / 858  & \checkmark &\\
     & && &  & & & &  & & & UNOS Transplant & NA / 23,706 &  \checkmark* &\\
        &&& &  & & & &  & & & MAGGIC  & NA / 30,389 & \checkmark* & \\
\midrule
\multirow{2}*{\cite{beaulieu2019privacy}} & \multirow{2}*{2019} & \multirow{2}*{AC-GAN} & \multirow{1}*{Generating Differntially private GAN via discriminator clipping} &  \multirow{2}*{\checkmark} &  \multirow{2}*{\xmark}  & \multirow{2}*{\xmark}  &  \multirow{2}*{\checkmark}&   \multirow{2}*{\checkmark}& \multirow{2}*{\checkmark} &\multirow{2}*{\checkmark} & SPRINT & 6,502 / NA & \checkmark* & \multirow{2}*{\checkmark} \\
  & & &  for tabular EHRs & & &  &  &  &  && MIMIC-III & 8,260 / NA & \checkmark* &\\

\midrule
\multirow{1}*{\cite{wang2020part}} & \multirow{1}*{2020} & \multirow{1}*{PART-GANs} & Improving private GAN training of time-series EHRs &\multirow{1}*{\checkmark} & \multirow{1}*{\xmark}& \multirow{1}*{\checkmark} & \multirow{1}*{\xmark}& \multirow{1}*{\checkmark} &  \multirow{1}*{\xmark} & \multirow{1}*{\xmark} & Philips eICU & 200,000 /  224,026,866 & \checkmark* & \multirow{1}*{\xmark}\\
\midrule
\multirow{2}*{\cite{yoon2020anonymization}} & \multirow{2}*{2020} & \multirow{2}*{ADS-GAN} & \multirow{1}*{Anonymizing generated tabular EHR data while}& \multirow{2}*{\checkmark} &\multirow{2}*{\xmark} & \multirow{2}*{\checkmark}& \multirow{2}*{\checkmark} & \multirow{2}*{\checkmark}  & \multirow{2}*{\checkmark}&\multirow{2}*{\checkmark} & MAGGIC (RCT data) & 30,389 / NA & \multirow{1}*{\checkmark*}  & \multirow{2}*{\checkmark} \\
     && & minimizing patient identifiability & & & &&  & & & 3 UNOS Transplant datasets & 23,706-56,822 / NA & \checkmark* &\\
 \midrule

 \multirow{2}*{\cite{yale2020generation}} & \multirow{2}*{2020} & \multirow{2}*{HealthGAN} & \multirow{1}*{Improved End-to-End privacy-preserving WGAN-GP with a } & \multirow{2}*{\checkmark} &  \multirow{2}*{\xmark} & \multirow{2}*{\checkmark} & \multirow{2}*{\xmark} & \multirow{2}*{\checkmark} & \multirow{2}*{\checkmark} &  \multirow{2}*{\checkmark}& \multirow{2}*{MIMIC-III} & \multirow{2}*{ NA / 27,000} & \multirow{2}*{\checkmark*}  &  \multirow{2}*{\checkmark}\\
      & & & focus on privacy \& resemblance metrics & &  && &   & &&& &\\
 \midrule
\multirow{2}*{\cite{indhumathi2021healthcare}} & \multirow{2}*{2021} & \multirow{2}*{HCGAN} & Improving robustness to privacy attacks by training Cramér&\multirow{2}*{\xmark} & \multirow{2}*{\xmark}& \multirow{2}*{\xmark} & \multirow{2}*{\xmark}& \multirow{2}*{\checkmark} &  \multirow{2}*{\checkmark}& \multirow{2}*{\xmark} & UCI Breast dataset & NA / 699 & \checkmark & \multirow{2}*{\xmark}\\
  & & & GANs for tabular EHRs & &  & &   && &&  Texas Hospital Data & NA / 186,976 & \xmark\\


 

\bottomrule
\end{tabular}

    \begin{tablenotes}
      \item [1] The included evaluation components are (DWS): Dimension-wise Similarity, (LDS): Latent Distribution Similarity, (JDS): Joint Distribution Similarity, (IDRS): Inter-dimensional Relationship Similarity, (PP): Privacy Preservation, and (DU) Data Utility, (Qual) Qualitative Evaluation, which are explained in details in section~\ref{eval}.
      \item [2] The dataset size is reported in the format of (N/R) where N refers the number of patients an R refers to number of records, reported in each of the works.
      \item [3] The symbol \checkmark* refers to data sources that can be accessed after going through an application process
     \end{tablenotes}
  \end{threeparttable}
  }
    \label{tab:EHR_GAN}
\end{table*}

\subsection{\textbf{Generation of Diverse Types of EHRs}}
\label{data_gen}
In the following subsections, we describe GAN-based works that generated different types of EHR data, tabular and  time-series, in sections~\ref{tab}, and~\ref{time}, respectivly. We also review papers that attempted to explore heterogeneity aspects in either tabular or timeseries EHRs, in section~\ref{heterogenous}.
\subsubsection{Generating Tabular EHRs}
\label{tab}
The early GANs for EHRs works focused on generating structured discrete tabular EHRs such as diagnosis and billing ICD codes. For example, \textit{medGAN} was one of the first GANs architectures to address the incompatibility of the original GANs to generate tabular EHRs with binary or discrete count features~\cite{choi2017generating}. The authors' model incorporated an autoencoder to learn the salient features of discrete variables in tabular EHRs, which assists GANs in learning the distribution of multi-label discrete binary and count features. 
Building on the success of \textit{medGAN} for generating discrete data, \textit{medWGAN} and \textit{medBGAN} were proposed based on Wasserstein GAN with gradient penalty \textit{(WGAN-GP)}~\cite{gulrajani2017improved} and boundary-seeking GANs \textit{(BGAN)}~\cite{hjelm2017boundary}, respectively. The authors' major contribution was in the area of improving the quality of generated data of that generated by the original~\textit{medGAN}~\cite{baowaly2019synthesizing}.
In \textit{MC-medGAN}, the authors proposed adaptations to medGAN to allow for better representation of multi-categorical data \cite{camino2018generating, goncalves2020generation}. To achieve this aim, the authors used a gumbel-softmax activation function to enable backpropagating for random samples of discrete variables, which has notable improvements for multi-categorical features~\cite{jang2016categorical}. 
Other researchers focused on improving the capturing of local correlations in tabular EHRs by proposed Correlation Capturing GAN \textit{(CorGAN)}~\cite{torfi2020cor}.~\textit{CorGAN} combined Convolutional GAN and Convolutional Autoencoders (CAs) to capture the local correlation between features in both discrete and continuous data. 
More recent works focused on improving the training stability, such as the work proposed in EMR Wasserstein GAN \textit{(EMR-WGAN)}. The authors removed the autoencoder which was inherited from medGAN to account for discrete features, applied a filtering strategy to enhance GANs training for low-prevalence clinical concepts~\cite{zhang2020ensuring}. With the new changes, EMR-WGAN was able to generate high-fidelity data with reduced noise, improved stability in GANs for EHRs training, and modeling~\cite{zhang2020ensuring}.

\subsubsection{Generating Time-series EHRs}
\label{time}
While it is useful to generate tabular EHR data that presents  patients' state at a single timepoint, tabular data does not capture the dynamics and changes effectively compared with time-series data, in which variables are recorded along a series of timepoints. 
To address this issue, a framework for Synthetic Temporal EHR Generation~\textit{(SynTEG)} was recently presented,  where the authors focused on generating timestamped diagnostic events (ICD-codes)~\cite{zhang2021synteg}. Their architecture approaches the problem in a two-stage approach. The first stage sequentially extracts temporal patterns from visits and adopts a self-attention layer~\cite{vaswani2017attention}. The second stage generates data conditioned on the learned patterns using Wasserstein GAN~\cite{arjovsky2017wasserstein1}.
In a similar application, the authors proposed to synthesize sequences of EHRs from patients' chronological visits by using the dual adversarial autoencoder \textit{DAAE} along with two GANs components \cite{lee2020generating}. By utilizing the recurrent autoencoder-based generators, DualAAE can synthesize sequences of set-valued medical records such as diagnosis ICD-codes.
Another GAN adaptation for continuous time-series EHRs was that of \cite{yahi2017generative}, whose work generated time-series drug laboratory effects (DLEs) trajectories. Their work has many applications for monitoring patients after exposure to interventions, which can prevent adverse drug reactions~\cite{tatonetti2011detecting}.
In ~\cite{esteban2017real}, the authors worked on a model to generate continuous time-series EHR data using Recurrent GANs \textit{RGAN}, and its conditional generative version \textit{RCGAN}. Recurrent neural networks, Long Short-Term Memory (LSTM) \cite{hochreiter1997long}, were use for both the generator and discriminator of RCGAN, which are commonly used for sequential data tasks \cite{hochreiter1997long, manaswi2018rnn}. 
Motivated by the clinical practice of dosage adjustments based on patient state and that both components have a mutual influence on each other,~\cite{Scgan} developed Sequentially Coupled GAN (\textit{SC-GAN}). Their model has two distinct LSTM-based generators that coordinate the generation of patient state and medication dosage data. The output of the patient-state generator is fed to the dosage generator, which mimics the clinical practice of assigning dosage based on the patient status~\cite{Scgan}. 

\subsubsection{\textbf{Generating Heterogeneous EHRs}}
\label{heterogenous}
To mimic the heterogeneous nature of EHRs which include various types (including demographic information, ICD codes, vital sign time-series, etc.), developing GANs that target synthesizing mixed-type EHRs and capturing the dependencies between various features is of vital importance.
In~\cite{chin2019generation}, the authors used \textit{WGAN} to generate discrete tabular EHR data containing both administrative and diagnostic data, which they referred to as heterogeneous EHRs. 

In parallel,~\cite{yan2020generating} developed a model to account for constraints and preserve relationships across generated heterogeneous tabular EHRs that combined binary, categorical and continuous values. To do so, the authors incorporated penalization for the violation during GAN training~\cite{yan2020generating}. 
To simultaneously generate continuous-valued and discrete-valued time-series EHRs, GANs for synthesizing Mixed-type longitudinal EHR data \textit{(EHR-M-GAN)} was lately proposed~\cite{li2021generating}. The authors utilized a dual variational autoencoder to generate a shared latent space representation of mixed EHR types. In addition, a sequentially coupled generator implemented with bilateral LSTM was adopted during data generation to capture the temporal correlations between heterogeneous types of EHRs. 


\subsection{\textbf{Semi-Supervised Learning and Data Augmentation}}
\label{semi-supervised}

It is often the case in healthcare datasets that different outcome classes are not equally prevalent, as seen in mortality and rare disease prevalence; this issue is referred to as class imbalance in the machine learning domain~\cite{japkowicz2002class}. Another commonly seen issue is the absence of labels for some samples, which is referred to as unlabelled samples. 
Learning from both labelled and unlabelled data gained increasing attention in the machine learning community,  where semi-supervised learning (SSL) approaches such as classification and clustering have proven to be effective in various applications~\cite{zhu2009introduction}.

Some researchers extended the GANs' role for SSL problems by forcing the GANs to output class labels for unlabelled samples~\cite{salimans2016improved, odena2016semi}. In their proposed setup, a GAN-based model is trained on a dataset with the samples belonging to one of K classes, where there's a high percentage of unlabelled samples. Then, the discriminator's role is adjusted to predict which of K+1 classes the samples belong to, where an extra class refers to the synthetic samples~\cite{salimans2016improved, odena2016semi}. The extension of the discriminator's role to predict classes opened the door for many applications with a high prevalence of unlabelled samples, such as the case of rare diseases where misdiagnosed or delayed diagnosis is common~\cite{knight2006common}. The work proposed by~\cite{li2018semi}  extended the discriminator's goal to finding the class assignment of real EHR samples to be able to detect rare diseases in a majority unlabelled tabular dataset. In addition, the authors used a modified loss for their generator, where the objective is to generate samples with minimal divergence from the target distribution. This objective is achieved by over-representing samples with low densities in the original distribution, referred to as ``complement samples'' as initially proposed by~\cite{dai2017good}.  Based on the success of~\cite{li2018semi}, the authors extended the work of GANs for SSL for predicting rare diseases to be compatible with longitudinal data~\cite{yu2019rare}. The main modification to the GANs models was the usage of RNNs for both the generator and discriminator architecture, which allowed for time-series generation.

GAN-based data augmentation methods have been proposed to mitigate imbalanced and unlabelled data. In such cases, generated data from a specific class can be used in conjunction with the real data to improve model performance, generalizability and decrease over-fitting~\cite{chaitanya2019semi}. Data augmentation can be beneficial when the target dataset has highly unlabelled points or is severely imbalanced, as seen in semi-supervised learning applications. For instance,~\cite{che2017boosting}, modified the original GANs and proposed \textit{ehrGAN} to learn the transition distribution of the samples by using a generator with variational contrastive divergence~\cite{zhai2016generative}. \textit{ehrGAN} is then used as a part of the loss function of a semi-supervised learning GANs framework \textit{SSL-GAN} to augment the training data in a semi-supervised learning manner for sequences of diagnosis codes. By learning the transition distribution of real samples,  rich structures of the data manifold around true examples are utilized in SSL-GAN to improve performance. 

In a similar application, \cite{yang2019gan} simultaneously addressed the problems of both the unlabelled and unbalanced data by using a GAN-based approach. The authors presented a framework in which the GAN takes labelled data as inputs and uses it to generate new samples. The generated labelled data are then used to train two independent classifiers to predict sample labels. Next, the authors used the classifiers' predictions to assign pseudo-labels for unlabelled samples. Samples with the same pseudo-label predictions from both classifiers are added to the labelled set. The authors then use GANs again to generate new samples in an attempt to re-balance the minority class labels~\cite{yang2019gan}. The final augmented dataset was used to train a classifier that achieved superior performance on various benchmark datasets. It is worth noting that in many of the semi-supervised uses of GANs, the generated data distribution does not need to match the real data distribution since the objective might be to over-represent the minority class~\cite{dai2017good}.  

\subsection{\textbf{Imputation of Missingness}}
\label{missingGAN}
Handling missing data remains one of the major challenges when dealing with EHRs, where data can be highly missing for various reasons. Using incomplete data for training machine learning algorithms can harm their performance, especially in cases where algorithms may not be robust to missingness~\cite{marlin2008missing}. Missing data is usually regarded as one of the following depending on the missingness pattern, missing completely at random (MCAR), missing at random (MAR), and not missing at random (NMAR)~\cite{little2019statistical}. In the healthcare domain, missingness can come in any of the three types, depending on the underlying missingness cause. Examples of healthcare-related causes of missing data in EHRs include,  data recording errors and machine failure, irregular sampling and inconsistent medical visits~\cite{kreindler2006effects}, unmeasured lab tests due to the lack of medical need~\cite{duff2019frequency}, or even high cost and dangerous to acquire information such as invasive or radiology procedures~\cite{bulas2013benefits,kim2017dangers} and other factors related to patient severity and diagnosis~\cite{agniel2018biases}. 

GANs are naturally suitable for generative tasks not only generating completely new samples, but also generating missing values that can be used to impute the original samples. While most data imputation methodologies are often based on either parametric or non-parametric probability density estimation, GANs can perform data imputation without calculating a probability density first~\cite{yang2019categorical}. The first GAN-based missing data imputation paper had a focus on image completion~\cite{iizuka2017globally}. This work motivated a series of GAN-based imputation methods that are application-specific and tailored for various data types including medical data. For instance,~\cite{yoon2018gain} proposed the use of an adjusted version of the original GANs which they refer to as Generative Adversarial Imputation Nets \textit{(GAIN)}. In their work, the generator's role was adjusted to generate and accurately impute missing data. The discriminator's role, however, was adjusted to distinguish between original and imputed components, analogous to distinguishing between real and synthetic samples~\cite{yoon2018gain}. To increase the performance and the quality of the generated imputation data, the discriminator is also given additional information ``hints'', which reveals to the discriminator partial information about the missingness of the original sample. Their work focused on MCAR missingness in multiple tabular datasets.
The results of \textit{GAIN} were benchmarked against various data imputation methods such as MICE~\cite{van2011mice}, missForest~\cite{stekhoven2012missforest} and Expectation-maximization \cite{nelwamondo2007missing}.

Others were motivated by the high missingness in the commonly used EHR data such as the \textit{MIMIC III} dataset \cite{johnson2016mimic}, where missingness reached as high as 74\%~\cite{zhang2018medical}. In their work, \cite{zhang2018medical} combined the structure proposed by~\cite{yoon2018gain} with principles of Stackelberg competition in the domain of game theory~\cite{fiez2019convergence}. The main adaptations of \textit{GAIN} are in the use of multiple generators (followers), rather than one, which team up against the discriminator (leader). Their results showed that the \textit{Stackelberg-GAN} was able to capture complex data distributions and achieved high performance when compared with other state-of-the-art imputation methodologies. The authors evaluated their work on discrete, continuous, and categorical tabular EHRs~\cite{zhang2018medical}.

In a similar work, \cite{yang2019categorical} proposed a modification to \textit{GAIN} that focused on improving performance in generating categorical tabular EHR data. The authors hypothesized that the original GANs architecture and the one used by~\cite{yoon2018gain} is not optimal for categorical features due to the softmax function's ability to produce values between 0 and 1~\cite{yang2019categorical}. To address this,~\cite{yang2019categorical} introduced a fuzzy binary coding of categorical features, where values are encoded using real numbers between 0 and 1 to preserve the categorical information aspect of the data. To further improve \textit{GAIN} for mixed-type tabular EHRs,~\cite{camino2019improving} modified its model structure where the generator and discriminators had multiple inputs as well as multiple outputs~\cite{camino2019improving}. The major contributions focused on variable splitting and the usage of gumbel-softmax activation which accounts for categorical variables and their discrete distributions~\cite{jang2016categorical}. While most works focused on MCAR cases, the authors of Multiple Imputation via GANs \textit{(MI-GAN)} introduced an architecture that is theoretically supported for both MCAR and MAR cases. The authors combined ideas from both \textit{GAIN} and Multiple Imputation machine learning works to solve the problem of MAR block-wise pattern missingness where the missing probabilities depend on the observed values in the dataset~\cite{dai2021multiple}. The results showed superior performance with respect to other imputation models in terms of statistical inference and computational speed. 

Despite the outstanding results of \textit{GAIN} and its various adaptations, they are not directly applicable to time-series EHRs. To fill this gap, the authors of~\cite{luo2018multivariate}, proposed a GAN-based model that is implemented with a modified Gate Recurrent Unit (GRU)~\cite{cho2014learning} to model the temporal irregularity of the incomplete time series data, which they refer to as GRU for data
Imputation (GRUI) cell. The use of GRU instead of LSTM and other RNN variants is motivated by its compatibility with the irregular time lags and variations between two consecutive observations including those seen in data such as ICU EHRs~\cite{luo2018multivariate}. GAN with GRUI model performs the imputation in a two-stage approach. First, it trains a GAN model to generate complete time-series, and then it tries for each sample, to find the "noise" vector that is most similar to the original sample~\cite{luo2018multivariate}. Despite reporting state-of-the-art results for imputing time-series EHR data, the work of~\cite{luo2018multivariate}, has a major drawback in terms of training efficiency. Motivated by improving the efficiency of GAN for time-series imputation, \cite{luo2019e2gan} proposed an end-to-end GAN-based imputation model, referred to as  \textit{E\textsuperscript{2}GAN}. The proposed model preformed imputations in reduced the training time, with higher quality by adopting a compressing and reconstructing
strategy to circumvent the noise optimization stage in the GAN with GRUI ~\cite{luo2018multivariate}.
Recently,~\cite{gupta2021concurrent} presented a novel GANs architecture~\textit{Bi-GAN} to perform both imputations of missing values and prediction of future values in time-series EHR data. Both the generator and discriminator were bi-directional recurrent neural networks Bi-RNNs, which are suitable for time-series applications. In their work, the GAN-based model learns from all the observed samples to impute missing values and then learns to predict future values, by treating them as missing values~\cite{gupta2021concurrent}. This problem setup does not require a definition of prediction windows at training time, which motivates flexible predictive models which they refer to as “any-time prediction tool”~\cite{gupta2021concurrent}.

\subsection{\textbf{Treatment Effect Estimation}}
\label{ITEgan}
Estimating treatment effects is a complicated causal inference task with many data challenges, where the aim is to estimate the patient's response to a specific treatment. The major challenges in this field arise from missing counterfactual data, the unobserved outcomes of untaken treatments~\cite{hill2011bayesian}. In Randomized Control Trial (RCT) settings, patients in the treatment group are matched to those in the control group to compensate for missing counterfactuals. However, despite being the golden standard for various clinical applications, RCT-based treatment effect estimation suffers from multiple issues concerning their high cost~\cite{sanson2007limitations}, relatively small size~\cite{hayes1999simple},  ethical issues~\cite{goldstein2018ethical}, and short duration of followups which might miss out long-term effects of medications~\cite{black1996we}. A low-cost alternative to RCT data is the regularly collected EHR data. Specifically, longitudinal EHRs, which include diverse patient cohorts, long-term outcomes with no strict exclusion criteria, making EHRs  more representative of the patient population~\cite{black1996we,newsome2018estimating}. However, in EHR data, treatments are not assigned at random, and there's no clearly defined control group. Thus, estimating treatment effects from EHRs requires measures to control confounding effects and perform covariate adjustment~\cite{mcnamee2003confounding,rosenbaum2002covariance} to avoid selection bias.

The generative capabilities of GANs are a valuable option for various treatment effect estimation applications. In \cite{yoon2018ganite}, the authors made use of GANs' generative properties to generate counterfactual outcomes. In their novel design, GANs for inference of Individualized Treatment Effects~\textit{(GANITE)}, they considered counterfactual outcomes to be missing labels, similar to their earlier work in \cite{yoon2018gain}. \textit{GANITE} utilized a pair of GANs: one for counterfactual imputation and another for treatment effect estimation. In the first GAN, the generator's task is adjusted to generate missing counterfactual outcomes, while the discriminator's task is to tell the factual from the counterfactual outcomes. In the presence of counterfactual outcomes, a treatment effect estimation function can be predicted using traditional machine learning models. However, in \textit{GANITE}, the authors utilize another GAN to model treatment effect estimation by taking the output of the counterfactual GAN as input and generating a potential outcome vector with confidence intervals~\cite{yoon2018ganite}. While \textit{GANITE} focused on binary treatment,~\cite{chu2020treatment} focused on generating time-series post-treatment outcomes. The authors' work was motivated by the scarcity of paired pre- and post-treatment patient time-series data in settings such as ICU ventilation and vasopressors assignment. Their proposed model, Cycle Wasserstein Regression GAN \textit{CWR-GAN}, is a hybrid of several architectures; original GAN~\cite{goodfellow2014generative}, Wasserstein GAN~\cite{arjovsky2017wasserstein1}, and cycle-consistent GAN~\cite{zhu2017unpaired}. The authors of CWR-GAN, tested their model in regression-based tasks, and provided an alternative to the traditional uni-directional regression approaches, where unpaired data would be ignored during training~\cite{chu2020treatment}.

To extend GAN-based treatment estimations from binary to various kinds of treatment variables including categorical, and continuous,~\cite{ge2020conditional} applied modifications to \textit{GANITE}, which they named \textit{MGANITE}. Estimating continuous treatment is of high importance in applications involving dosage adjustment especially in oncology~\cite{menz2021barriers}. One of the main modifications was a mathematical adjustment to the loss function which takes a treatment assignment vector in both the counterfactual and ITE estimation blocks, to allow for simultaneous treatment effect estimation~\cite{ge2020conditional}. When using observational data where treatments are not randomly assigned, controlling the confounding factors, such as using propensity scores is essential~\cite{d1998propensity}. In~\cite{li2020continuous}, the authors propose a GAN-based model that generates a ``calibration'' distribution, one that eliminates associations between covariates and treatment assignment by a random perturbation process of the treatment variable. The generative capabilities of GANs are used to learn a weight vector that is used to adjust the distribution of observed data and construct the calibration data. The authors refer to their model as Generative Adversarial De-confounding (\textit{GAD})~\cite{li2020continuous}. 

Statistical approaches such as propensity score matching (PSM) are commonly used by classical treatment effect estimation works to balance the population's characteristics assigned either to an intervention or a control group~\cite{caliendo2008some}. However, despite their popularity, PSM approaches can lead to high reductions in sample sizes due to unmatched control samples~\cite{caliendo2008some}. Lately, a GAN-based propensity score synthetic augmentation matching model, \textit{PSSAM-GAN}, was proposed to mitigate the problem of sample size reduction using PSM approaches~\cite{ghosh2021propensity}. First, the authors matched their samples based on calculated propensity scores. Then, to be able to use unmatched samples, the authors used a GAN-based model to generate treatment matches for the unmatched control samples~\cite{ghosh2021propensity}. Finally, the original EHR data was augmented with the newly generated matched samples to be used for downstream treatment estimation tasks~\cite{ghosh2021propensity}.
\subsection{\textbf{Privacy Preservation}}
\label{privacy}
\begin{figure}[htpb!]
\centering
    \resizebox{0.6\linewidth}{!}{
    \includegraphics{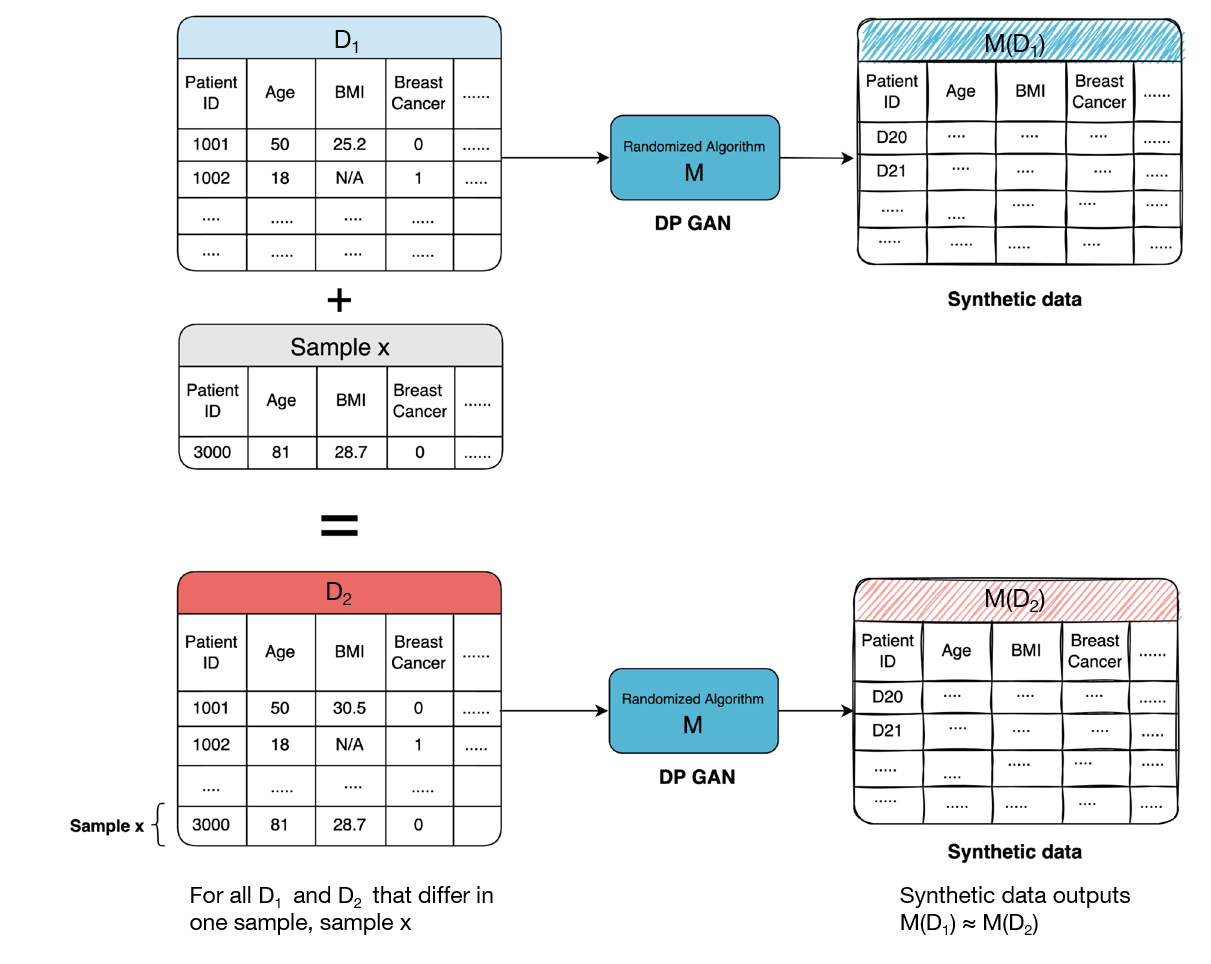}}
    \caption{GAN training with differential privacy guarantees. Real datasets D\textsubscript{1} and D\textsubscript{2} only differ in a single sample X. ${M}$, is the differentially private GAN model, that outputs ${M}$(D\textsubscript{1}), and ${M}$(D\textsubscript{2}) which at most have a difference of \textbf{e\textsuperscript{$\epsilon$}} }
    \label{fig:DP_gan}
\end{figure}

Privacy is a central theme in GAN development as it is a principal motivator for using generative models in healthcare applications. Even though GANs do not explicitly expose patient data, some works demonstrated the importance of improving the privacy-preservation of GANs, especially when dealing with sensitive information such as patient EHRs~\cite{mukherjee2020protecting}. In the field of privacy, there has been a wave of frameworks that apply theoretical guarantees to ensure the privacy of the data~\cite{usynin2021adversarial}. Notably, differential privacy is a theoretical guarantee that allows learning nothing about an individual while learning useful information about a population~\cite{dwork2014algorithmic}. Differential privacy is concerned with the impact of the presence or absence of a single record on the outcome of the computational tasks. Differential privacy, is defined as follows: 

A randomized algorithm A is $\epsilon\text{-}$differentially private if for any two datasets D\textsubscript{1} and D\textsubscript{2} that differ in a single point and for any subset of outputs S:
\begin{equation*} P(M(D_1)\in S) \leq e^\epsilon P(M(D_2) \in S). \end{equation*}
  where P is taken with respect to the randomness, ${M}(D_1)$ and ${M}(D_2)$ are the outputs of the ${M}$ for databases D\textsubscript{1} and D\textsubscript{2}, respectively~\cite{dwork2014algorithmic}. Based on this definition, there are many differentially private algorithms, any of which may be used to complete the same computational task under differential privacy guarantees~\cite{dwork2014algorithmic}. Differential privacy can be applied to GAN training, where M refers to the differntially private GANs as seen in Figure~\ref{fig:DP_gan}.

Motivated by improving privacy through providing theoretical guarantees for medical data, several works developed and evaluated differentially private GANs for EHRs generation applications. 
Namely, DPGAN~\cite{xie2018differentially} proposed GANs with differentially private guarantees by adding noise to the discriminator's gradients, which was inspired by moment accountant techniques~\cite{abadi2016deep}. Similarly,~\cite{jordon2018pate}, proposed a modification to the GAN training of the discriminator by using an adaptation of the differentially private framework, Private Aggregation of Teacher Ensembles (PATE)~\cite{papernot2016semi}. In PATE, multiple teacher models are independently trained on subsets of the data for a classification task. The final classification output is an aggregate of each of the teacher model's prediction~\cite{papernot2016semi}. Another differentially-private GANs for EHRs development was~\cite{beaulieu2019privacy}, where the authors limited the effect of a single participant on the training by clipping the norm of the discriminator's gradient combined with the addition of Gaussian noise. In a similar spirit,  the authors of \cite{wang2020part} proposed a data augmentation framework with differential private guarantees and model optimizations to improve the data utility without compromising the quality. The proposed framework, privacy-preserving Augmentation and Releasing scheme for Time series data via GAN (\textit{PART-GAN}) uses weight pruning and grouping, generator selecting, and denoising mechanisms for improving the quality in time-series data~\cite{wang2020part}.
Some works combined both theoretical and empirical evaluations to prove the privacy-preservation of the GAN model~\cite{indhumathi2021healthcare}. To avoid compromising the synthetic data fidelity, the authors applied partial differential privacy to the Quasi Identifier features; these features are then recombined with the other sensitive attributes. The authors then trained a GAN that relies on Cramér distance~\cite{bellemare2017cramer} between the joint distribution of the generated observation and real differentially private patient data using the combined feature set. The model was then tested for various adversarial attacks to support their theoretical guarantees~\cite{bellemare2017cramer}.

Despite the strong privacy guarantees of differential privacy, it has various technical limitations such as compromised data fidelity and utility. This motivated works to look for strong privacy-preserving alternatives. For example,~\cite{yoon2020anonymization} developed a WGAN-PB-based model which they refer to as anonymization through data synthesis using generative adversarial networks \textit{(ADS-GAN)}. In their work, the authors created a mathematical definition for “identifiability”, which was based on the probability of re-identification given the combination of all data on any individual patient~\cite{yoon2020anonymization}. In \textit{ADS-GAN}, the authors tested for the data quality, while maintaining the identifiability constraints. In a similar notion,~\cite{yale2020generation} worked on an end-to-end privacy-preserving GAN based on WGAN-PB, and proposed a quantitative privacy metric, privacy\textsubscript{loss} which is based on the balanced accuracy of the adversarial nearest-neighbors model. 


\section{\textbf{Evaluation of GANs for EHRs}}
\label{eval}

Despite the substantial attention given to theoretical and application-oriented GAN development gained over the past years, there is still no consensus on evaluation metrics or methodologies~\cite{theis2015note}. Evaluating the strengths and shortcomings of the model and synthetic data is vital for fair benchmarking and future research directions. For example, evaluating whether the GAN model is simply memorizing training examples or is missing important information and characteristics relating to data distribution is essential prior to using the synthetic data for downstream tasks. The evaluation of GAN models can take various directions all of which have different aims such as close approximation of data distribution, maintaining privacy, the utility for downstream machine learning tasks, and model performance. Evaluation methods described in the literature, including those seen in the papers presented in Table~\ref{tab:EHR_GAN} can be grouped into two groups (1) qualitative and (2) quantitative evaluation methods. In Table ~\ref{tab:metric}, we present a list of the different quantitative evaluation metrics and tests used in the reviewed papers in this work, along with the data types each metric was used to evaluate, and the references to the works that explain each of the metrics.

\begin{table}
    \centering
     \caption{Quantitative metrics and tests used for evaluating GANs for EHR models}
        \resizebox{0.7\linewidth}{!}{
           \begin{threeparttable}[b]
    \begin{tabular}{llcc}

    \toprule
          \multirow{2}*{\textbf{Reference}} &     \multirow{2}*{\textbf{Evaluation metric}} & \multicolumn{2}{c}{\textbf{Data Type}}\\
            &  & Tabular & Time-series\\
        \midrule
        \textbf{Dimension-wise Distribution Similarity} & &&\\
    
        \midrule
      \multirow{1}*{\cite{choi2017generating}} & \multirow{1}*{Dimension-Wise Probability}  & \checkmark 	&	\checkmark	\\
     \midrule
     \cite{baowaly2019synthesizing} & Dimension-Wise Average &\checkmark	&	NA\\
           \midrule

     \multirow{1}*{\cite{massey1951kolmogorov}} &\multirow{1}*{Kolmogorov–Smirnov (K-S) test } &  \checkmark	&	NA\\
     \midrule
        \cite{goncalves2020generation} & Support Coverage & \checkmark	&	NA \\
    \midrule

      \cite{belov2011distributions} & Kullback-Leibler Divergence (KLD)& \checkmark	&	NA \\
        \midrule
      
     \textbf{Latent Distribution Similarity} &&&\\
     \midrule
     \cite{zhang2020ensuring} & Latent Space Representation & \checkmark	&	NA\\
  
     \midrule
     
    \multirow{1}*{\cite{zhang2021synteg}} & \multirow{1}*{Weighted Latent Difference} & NA	&	\checkmark\\
    
    \midrule
   
    \cite{woo2009global} & Log-cluster & \checkmark	&	NA \\
    \midrule
        \textbf{Joint Distribution Similarity} &&&\\
   \midrule
      \cite{belov2011distributions} & Kullback-Leibler Divergence (KLD) & \checkmark	&	NA \\
        \midrule

     \cite{majtey2005jensen} & Jensen-Shannon Divergence (JSD) & \checkmark	&	NA\\
      \midrule
      \cite{ruschendorf1985wasserstein,vallender1974calculation} & Wasserstein Distance  & \checkmark	&	NA\\
            \midrule
    \cite{gretton2012kernel} & Maximum Mean Discrepancies (MMD) &NA	&	\checkmark\\
   
    \midrule

     \cite{salimans2016improved} & Inception Score (IS) & NA	&	\checkmark\\
     \midrule
     \cite{yan2020generating} & Cross-type Conditional Distribution & \checkmark	&	NA \\
      \midrule
     
     \cite{zhang2020ensuring} & First-Order Proximity & \checkmark	&	NA\\
          \midrule
      
     \cite{yoon2019time, zhang2018decoupled} & Discriminative score  & NA	&	\checkmark\\
     \midrule
    
             \textbf{Inter-dimensional relationship Similarity} & &&\\
      \midrule
        \multirow{1}*{\cite{choi2017generating}} & \multirow{1}*{Dimension-wise Prediction} & \checkmark	&	NA \\ 
     \midrule
   
    \cite{benesty2009pearson} & Pairwise Pearson Correlation &\checkmark	&	\checkmark\\
\midrule

     \cite{zhang2003association} & Association Rule Mining (ARM) & \checkmark	&	NA\\
    \midrule
    \cite{yan2020generating} & Frequent Association Rules (FAR) & \checkmark	&	NA\\
  \midrule

    \textbf{Privacy Preservation } &\\
    \midrule
    \cite{dwork2014algorithmic} & Differential Privacy Guarantees &\checkmark	&	\checkmark\\
    \midrule
    \cite{shokri2017membership} & Member Inference Attack & \checkmark	&	\checkmark \\
    \midrule
    
    \cite{matwin2015review} & Attribute Disclosure Attack & \checkmark	&	\checkmark\\
        \midrule
    \cite{fredrikson2015model} & Model Inversion Attack & NA	&	\checkmark \\
        \midrule
       \cite{yoon2020anonymization} & Identifiability & \checkmark	&	NA\\
    \midrule
    \cite{yale2020generation} & Privacy\textsubscript{loss} & \checkmark	&	NA\\
    \midrule
    \cite{weldon2021generation} & Exact-matches test & \checkmark	&	NA\\
      \midrule
    \textbf{Data Utility} &\\
    \midrule
    \cite{esteban2017real} & Train on Synthetic, Test on Real (TSTR) & \checkmark	&	\checkmark\\
      \midrule
      \cite{esteban2017real} & Train on Real, Test on Synthetic (TRTS) &\checkmark	&	\checkmark \\
    \midrule
    
     \cite{lee2020generating,zhang2021synteg} & Predictive Modeling, Forecast Analysis & \xmark 	&	\checkmark\\
     \midrule
     \cite{jordon2018pate,jordon2020synthetic} & Synthetic Ranking Agreement (SRA)  & \checkmark	&	NA\\
    \midrule
    \cite{Scgan} & Data Augmentation Test  & NA	&	\checkmark \\

      \bottomrule
    \end{tabular}
  \begin{tablenotes}
   \item [1] The works referenced in the first column of table~\ref{tab:metric}, refer to the papers that explain the respective evaluation metrics.
   \item [2] The \checkmark symbol refers to metrics that were utilized to evaluate synthetic data for the corresponding data type, while NA refers to those with no available validation in the literature.
     \end{tablenotes}
  \end{threeparttable}
  }
    \label{tab:metric}
\end{table}
\subsection{\textbf{Quantitative Evaluation}}

    \subsubsection{\textbf{Dimension-wise Distribution Similarity}}
   A major objective of generative models is generating data whose distribution highly resembles that of the real dataset. Many evaluation metrics have been proposed to quantitatively evaluate the distribution resemblance per feature or ``dimension''. For instance, Dimension-wise probability is a test that compares the probability distribution of each of the features in real and synthetic datasets. The comparison method varies depending on the structure and type of data. For example, the Bernoulli success probability or Pearson Chi-square test were used for binary features~\cite{choi2017generating,zhang2020ensuring,yan2020generating,xie2018differentially,yoon2020anonymization}, while in other works the Student T-test was used continuous variables~\cite{yoon2020anonymization}. A similar evaluation test Dimension-wise Average was introduced to account for discrete count variables such as disease or procedure codes. The test simply calculates the dimension-wise average and compares that of the real to the synthetic dataset~\cite{baowaly2019synthesizing}. Another commonly used test is the  Kolmogorov-Smirnov K–S test, which simply tests that two samples came from the same distribution~\cite{massey1951kolmogorov}. The test is based on a well-known statistical metric, which is calculated by finding the maximum absolute value of the differences in the cumulative distribution functions of the two compared samples as seen in~\cite{baowaly2019synthesizing}. Other works took less rigorous evaluation approaches by reporting the distributions and statistical values as mean and standard deviation of both the synthetic and real datasets~\cite{beaulieu2019privacy,wang2020part}. To measure the extent of variable distribution coverage in the synthetic data,~\cite{goncalves2020generation} used support coverage metric. In this metric, the ratio of the cardinalities of a variable’s support is calculated in the real and synthetic data. The final result aggregates the results overall variables, to measure the joint support coverage. While more commonly used to measure overall data divergence, some papers used divergence metrics such as the Kullback Leibler Divergence (KLD), which is also known as the relative entropy on the feature level, as seen in~\cite{goncalves2020generation}. KLD is used for many applications to calculate a score, or distance, that quantifies the divergence of one probability distribution from another~\cite{belov2011distributions}. KLD is seen for many applications including  Gaussian Mixture Models, and t-distributed stochastic neighbor embedding.  By definition, KLD is defined to be:
   
     \begin{align*} \boldsymbol{KLD}(\mathcal {P, Q})=\int_{\mathcal{X}} \mathcal P(x) \log \frac{\mathcal P(x)}{\mathcal Q(x)} dx \end{align*}
     for distributions $\mathcal{P}$ and $\mathcal{Q}$~\cite{belov2011distributions}.
     
    \subsubsection{\textbf{Latent Distribution Similarity}}
    Building on the intuition that a good GAN model generates synthetic data that captures lower-level relationships even in the latent space, several works evaluated the latent distribution similarity between the real and synthetic datasets. For example,  \cite{zhang2020ensuring,yan2020generating} used a Latent Space Representation (LSR) test, where real and synthetic samples are projected into the latent space by utilizing a $\beta$ variational autoencoder~\cite{higgins2016beta}. After obtaining the projection in latent space, the dimensional mean of the distribution variance of each of the latent features is calculated in the synthetic data and compared to that of the real counterpart. A smaller distance or difference corresponds to a higher resemblance. This metric becomes of higher relevance when considering applications where interpretability is an integral component. Latent space evaluation metrics were also used by~\cite{zhang2021synteg}, where the authors calculated a weighted K–S average across all latent features. The latent space presentation and weights were arrived at by applying Singular Value Decomposition~\cite{klema1980singular} which yielded singular vectors and the corresponding singular values (weights) for each of the features.  The calculated weighted averages for the synthetic and real data were compared to test for similarity in the latent space representation. Another way to measure the similarity in the latent space is by using unsupervised learning approaches such as the log-cluster metric~\cite{woo2009global} as seen in~\cite{goncalves2020generation}. To measure the similarity of the underlying latent structure of the real and synthetic data, both datasets are merged and clustered using K-means clustering. Disparities of cluster membership of real samples versus synthetic samples are indicative of latent representation divergence~\cite{goncalves2020generation}.  
     
    \subsubsection{\textbf{Joint Distribution Similarity}}
    Preserving the real data distribution is a major aspect of evaluating the GAN quality. Aside from evaluating the distribution at the individual feature level,  synthetic data needs to be evaluated in terms of preserving the joint distribution of the real data. Joint distribution is usually evaluated by calculating one of the distance metrics such as KLD~\cite{belov2011distributions} as seen in~\cite{ge2020conditional}.  However, one of the major drawbacks of KLD, is that it is not symmetrical, where  KLD$(\mathcal P, \mathcal Q) \neq$ KLD$(\mathcal{\mathcal Q,P)}$. To overcome this, GAN-based models can be more accurately evaluated using Jensen-Shannon Divergence (JSD)~\cite{majtey2005jensen}. The definition of JSD builds on KLD; which is defined as follows:
    \begin{align*} &\boldsymbol{JSD}(\mathcal P, \mathcal Q ) = \frac{1}{2} KLD(\mathcal P, \mathcal M) + \frac{1}{2}KLD(\mathcal {\mathcal Q}, \mathcal M)
    \end{align*}
    where $\mathcal M$ is the average distribution with density $ 1/2 * (\mathcal P + {\mathcal Q}) $, for distributions $\mathcal{P}$ and $\mathcal{Q}$~\cite{nielsen2020generalization}.
    
 Results using JSD are symmetrized and smooth, which explains its usage in training critic of many GAN applications including the original GAN architecture~\cite{goodfellow2014generative} as well as in the evaluation of some of the GANs for EHR applications~\cite{yoon2020anonymization}. Another KLD-based metric is the inception score (IS), which was introduced by~\cite{salimans2016improved}, and is commonly used in many imaging applications. Despite capturing the quality and diversity of the data, IS is highly sensitive to noise, and thus it is rarely used in evaluating GANs for EHRs models~\cite{wang2020part}. 
    
Another joint distribution metric used is based on the Wasserstein distance, also referred to as Earth Mover’s Distance (EMD) which informally measures the minimum mass displacement to transform one distribution into the other~\cite{vallender1974calculation}. Even though this is a metric used for evaluating the joint-distribution similarity in the synthetic data, it is more often used in training loss function as seen in the well-known Wasserstein GAN which was introduced to overcome overfitting and mode collapse issues in GAN training~\cite{arjovsky2017wasserstein1,arjovsky2017towards}. The Wasserstein distance for $P$ and $Q$  distributions over $\mathcal{X}$ is defined as:

\begin{align*}
\boldsymbol {WD}(P, Q)=\inf _{\gamma \in \Gamma} \int_{\mathcal{X} \times \mathcal{X}}\|x-y\|_{2} d \gamma(x, y)
\end{align*}

where $\Gamma$ is the set of all possible joints on $\mathcal{X} \times \mathcal{X}$ that have marginals $P$ and $Q$~\cite{vallender1974calculation}.

    The usage of WD as a training critic has been particularly seen in many GANs for EHRs works reviewed in this paper~\cite{baowaly2019synthesizing,zhang2020ensuring,yoon2020anonymization,weldon2021generation}, while fewer works used it as an evaluation metric for joint similarity~\cite{yoon2020anonymization}. One major drawback of the WD is that it tends to be intractable in high dimensions, as well as its high computational complexity, and  biased sample gradients~\cite{karras2017progressive,arora2017generalization,bellemare2017cramer}. Another commonly used quantitative evaluation metric is Maximum Mean Discrepancies (MMD), which was first introduced in 2012 as a kernel two-sample test~\cite{gretton2012kernel}. MMD measures the dissimilarity between two probability distributions and uses samples drawn independently from each distribution.~\cite{gretton2012kernel}. MMD relies on the idea of representing distances between the compared distributions as differences of feature embeddings, mapped using Reproducing Kernel Hilbert Space (RKHS)~\cite{gretton2012kernel,berlinet2011reproducing}.  More formally, MMD between two distributions $P$ and $Q$ over $\mathcal{X}$ in in the the RKHS Kernel $\mathcal{H}_{k}$ is: 

    \begin{align*}
    \boldsymbol{MMD}_{k}^{2}(P, Q):=\mathbb{E}_{x, x^{\prime}}\left[k\left(x, x^{\prime}\right)\right]+\mathbb{E}_{y, y^{\prime}}\left[k\left(y, y^{\prime}\right)\right]-2 \mathbb{E}_{x, y}[k(x, y)]
    \end{align*}

    where $x, x^{\prime} \stackrel{i i d}{\sim} P$ and $y, y^{\prime} \stackrel{i i d}{\sim} Q $~\cite{gretton2012kernel}.

     Some works proposed novel joint distribution similarity tests that focus on the overall preservation of conditional distribution. For example,  Cross-type Conditional Distribution (CCD)~\cite{yan2020generating}  metric evaluates if the synthetic data maintains the distribution of one data type conditioned on another. The conditional distribution is quantified in terms of the mean and standard deviation and then compared between synthetic and real datasets. Fist-Order Proximity (FOP) is another metric introduced by~\cite{zhang2020ensuring}  measures the similarity of the structural associations between the real and generated datasets. To do so, an undirected graph is generated in which the weight of an edge between categorical features, such as diagnosis codes,  corresponds to their co-occurrence frequency in the population. The difference in FOP between the synthetic data and real data is calculated and used as a metric of preserving the associations. Other researchers evaluated  joint distribution similarity using  unsupervised clustering-based evaluations can be employed as seen~\cite{yahi2017generative}. Similarly, an unsupervised-based evaluation was introduced by~\cite{yale2020generation}, where the adversarial accuracy of a clustering model is used to capture resemblance loss of the GAN model, which the authors refer to as Train and Test resemblance losses. 
     
     Other than the aforementioned unsupervised-based evaluation, some authors leveraged an additional supervised task to quantify GANs' performance. 
     A binary classifier (a post-hoc discriminator) is trained to discriminate between the synthetic samples and the held-out real samples. The performance of the model, the discriminative score, is used to quantify the synthetic data's resemblance to the real data  without calculating statistical distances \cite{ li2021generating, lee2020generating}.

    \subsubsection{\textbf{Inter-dimensional Relationship Similarity}}
    Other than evaluating for the dimension-wise and joint similarity, it is important to also assess the synthetic data's preservation of inter-dimensional relationships and correlation between features. Several works used the  Dimension-wise Prediction test introduced by~\cite{choi2017generating}. This test iteratively chooses a feature and assigns it as a label, and treats the rest of the features as inputs. Two classifiers are trained, where one is trained on real data and the other is trained on the synthetic data to predict the selected label~\cite{choi2017generating,torfi2020cor}. The performance for each of the trained classifiers per feature is then compared, the assumption is that the closer the performance of pairs, the better the quality and inter-dimensional relationship similarity of the synthetic dataset~\cite{baowaly2019synthesizing,choi2017generating,torfi2020cor,yan2020generating, zhang2020ensuring,xie2018differentially}. The trained classifiers are usually logistic regression models~\cite{choi2017generating,torfi2020cor}, but at other times different classifiers such as support vector machine (SVM) and random forest were used~\cite{baowaly2019synthesizing}. Other works conducted inter-dimensional correlation evaluations such as Pearson Coefficient Correlation matrices comparisons for both real and synthetic data~\cite{beaulieu2019privacy,yoon2020anonymization,goncalves2020generation,Scgan}. The resulting mean vector and covariance matrices are compared to evaluate the resulting dataset for preserving inter-dimensional correlations and relationships.
    
    Association Rule Mining (ARM), is commonly used  in clinical data-mining applications. ARM models are used to identify meaningful patterns rules among clinical concepts~\cite{yadav2018mining,shin2010diagnostic}.  The GANs' ability to preserve the rules identified in the real set was evaluated by using a machine learning ARM model to identify association rules and compare those derived from the real to those of the synthetic~~\cite{baowaly2019synthesizing}.  Other authors introduced Frequent Association rules (FAR)~\cite{yan2020generating}, which utilizes the theoretical bases of ARM. FAR checks for both support and confidence, where supports represent how frequently the condition set appears in the dataset, whereas confidence is an indication of how often a condition rule is true~\cite{yadav2018mining}. After applying ARM, the proportion of the association rules that appear in both the real synthetic data are compared and then reported in terms of classification performance metrics such as precision and recall~\cite{yadav2018mining}.

    \subsubsection{\textbf{Privacy Preservation}}
\begin{figure*}[htpb!]
    \centering
    \resizebox{0.97\textwidth}{!}{
    \includegraphics{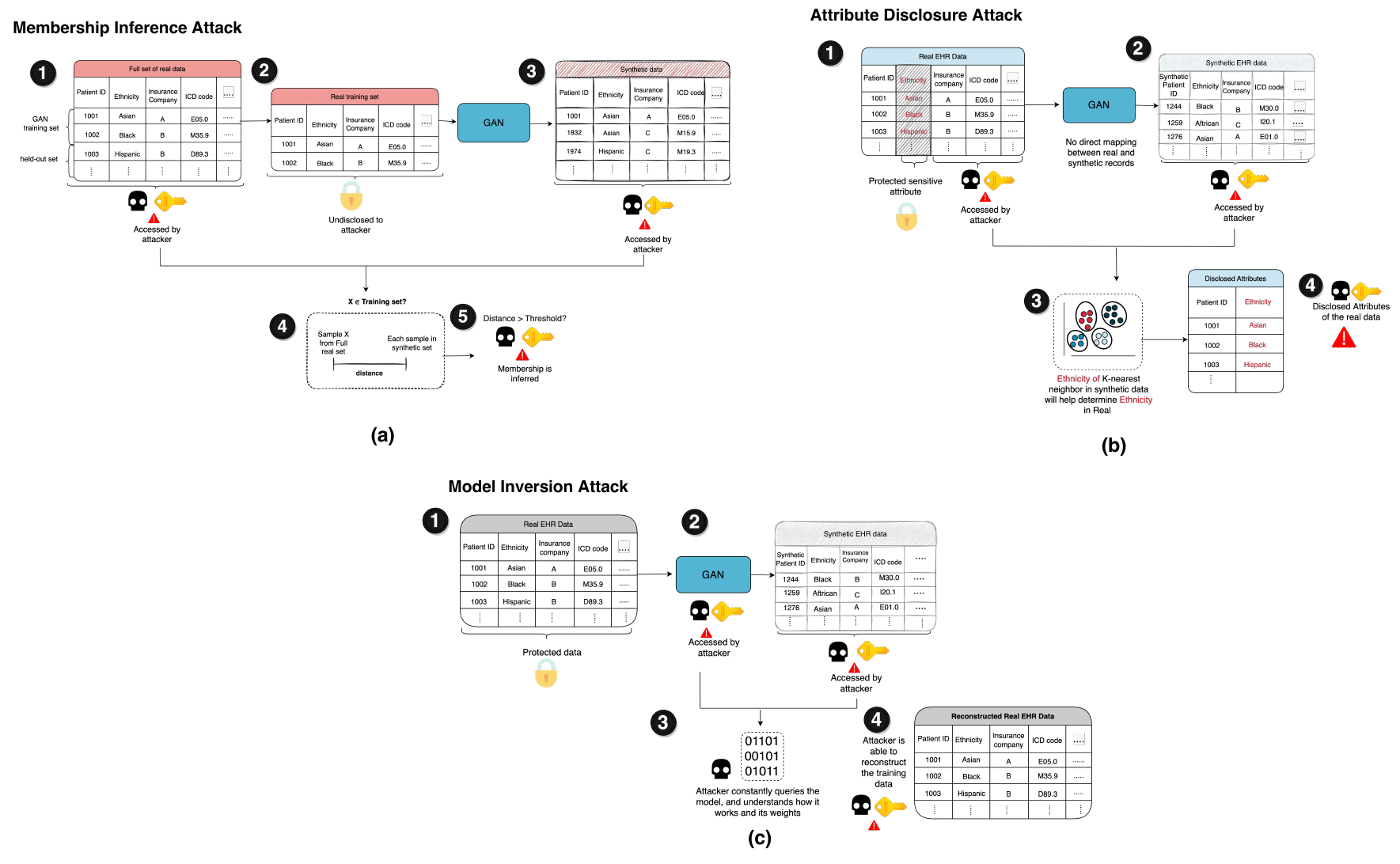}}
    \caption{The major types of adversarial attacks used for empirical evaluations of GAN models. (a) Membership Inference Attack (b) Attribute Disclosure Attack (c) Model Inversion Attack.}
    \label{fig:privacy}
\end{figure*}

    Evaluating the quality, and fidelity of the synthetic data is essential. However, to ensure safe usage of the resulting synthetic data, there's also a need to make sure that patients' privacy is not compromised. As there is no universally accepted standard definition for privacy~\cite{krumm2018ubiquitous}, the works reviewed in this paper dealt with privacy evaluation in a wide range of ways. Theoretical privacy guarantees such as differential privacy have been used in many of the GANs for EHRs works, as seen in~\cite{esteban2017real,jordon2018pate,chin2019generation,beaulieu2019privacy,wang2020part,xie2018differentially}. With the strict differential privacy's guarantees that neatly confirm privacy preservation, such works generally did not undertake further information leakage evaluation. While such approaches might seem ideal, differential privacy might lead to compromised data and utility preservation~\cite{domingo2021limits} as seen in~\cite{esteban2017real,chin2019generation}. An alternative to theoretical guarantees is the empirical evaluation of the robustness to well-studied attacks. The attacks evaluated in the reviewed papers include (a) membership inference attacks, (b) attribute disclosure attacks, and (c) model inversion attacks.
    First, membership inference (MI) attacks, where it is assumed that the attacker has access to the records of a set of real patient records, and attempts to determine if anyone from the real patients is in the training set of the GAN model~\cite{shokri2017membership}. To test for MI scenarios, a distance metric is calculated between each record in both the real and synthetic datasets. A threshold is then chosen as a cutoff, such that any record from the synthetic data with a distance less than the threshold is considered from the training set. Some works calculated this distance using hamming distance~\cite{choi2017generating,yan2020generating,zhang2020ensuring,goncalves2020generation}, while others used cosine similarity~\cite{torfi2020cor,weldon2021generation}. The performance is then reported in terms of precision, and recall to quantify the GANs' robustness to MI attacks. In other instances, a model is used to estimate the likelihood for a given record referred to as perplexity and then report metrics such as R\textsuperscript{2} and KLD are used to estimate the extent of distribution similarity as a proxy log likelihood~\cite{zhang2021synteg}. An overview of a sample MI attack is shown in Figure~\ref{fig:privacy} (a).
    
    The second type of adversarial scenarios is attribute disclosure (AD) attacks which occurs when an attacker can infer additional attributes about a patient by knowing a subset of other attributes about the same patient~\cite{matwin2015review}. To simulate this scenario, a random percentage of the real training set is sampled as well as a random set of features to be those disclosed to the attacker~\cite{choi2017generating}. A voting-based k-nearest neighbor clustering classification is utilized to estimate the values of the known features and then performance metrics in terms of precision and recall are reported as seen in~\cite{choi2017generating,yan2020generating,zhang2020ensuring,goncalves2020generation}. Some works extended this simulation by assuming the worst-case scenario where the attacker also has prior statistical knowledge about the undisclosed  features~\cite{zhang2021synteg}.  An example attribute disclosure attack is shown in Figure~\ref{fig:privacy} (b). The other type of attacks, namely model inversion refers to the scenario where an attacker aims to reconstruct the training data by their ability to constantly query the model~\cite{fredrikson2015model}, as shown in  Figure~\ref{fig:privacy} (c). This kind of attack was not frequently used in GANs for EHRs evaluation~\cite{indhumathi2021healthcare}, due to its replication complexity. 
    The aforementioned attacks can be implemented under two different scenarios against the generative models, either black-box or white-box setting~\cite{hayes2019logan}. In a white-box setting, the attacker has full access to the target model, including the architecture and weights of a trained network. While in a black-box setting, the attacker is only able to make queries to the target model and has no knowledge of its internal parameters as implemented in~\cite{li2021generating}.
    
    Some papers also developed a mathematical definition of privacy, such as identifiability which refers to the probability to re-identify samples included in the training~\cite{yoon2020anonymization}. Similarly,~\cite{yale2020generation} proposed an unsupervised adversarial privacy-based privacy-loss metric to quantify the extent of privacy preservation. Lastly,  simple evaluations such as Exact-Matches test were applied to check for the presence of exact duplicates of the training data in the synthetic data~\cite{weldon2021generation}.

    \subsubsection{\textbf{Data Utility}}
    High-quality synthetic data is a valuable asset for various research purposes as seen in section~\ref{GAN}. Evaluating the synthetic data in terms of its utility is a practice that has been adopted by many of the works reviewed in Table~\ref{tab:EHR_GAN}. One of the earlier machine learning utility testing frameworks was proposed by~\cite{esteban2017real}, which is Train on Synthetic Test on Real (TSTR). As the name implies, a machine learning model is trained on synthetic data and then tested on held-out real data. Similarly, Train on Real, Test on Synthetic (TRTS) was also proposed by ~\cite{esteban2017real}, as a reverse case of TSTR. When evaluating TSTR, the results show the utility of synthetic data when used for model buildings and conducting analysis; however, the model is applied to real data. On the other hand, TRTS could potentially supplement the performance of a model that is trained and tested on real data, with results on a synthetic dataset based on a dataset from a different source,  where access to the second dataset might not be feasible.
    The framework is flexible and can be used on any task-based machine learning application such as supervised classification~\cite{esteban2017real} where classification metrics such as F1 score, accuracy, and precision can be reported on both the synthetic and real datasets \cite{hossin2015review}. Other works assessed TSTR for supervised regression~\cite{zhang2021synteg,che2017boosting,chin2019generation}, where metrics such as Area Under Receiving Operator Curve (AUROC), and Area Under Precision-Recall Curve (AUPRC)~\cite{hossin2015review} were reported. Semi-supervised learning works focusing on mitigating data imbalance issues evaluated the utility of synthetic data for machine learning tasks for the same purpose~\cite{che2017boosting,li2018semi}. 
    Time-series specific supervised learning evaluations were be applied to generative tasks to evaluate the preservation of temporal dynamics~\cite{lee2020generating,zhang2021synteg}. The same temporal-related supervised task on both the real and synthetic datasets, such as predicting the top-N ICD codes in patient's next visit \cite{lee2020generating}, or forecasting patient's future diagnosis \cite{zhang2021synteg},  which were referred to as predictive modeling performance or forecast analysis, respectively. A similar performance of the models on both the synthetic and real dataset is indicative of the GANs' ability to preserve characteristics and utility of the real data.
    
   Despite their wide use, TSTR, TRTS and other data utility evaluations are sensitive to the model chosen for evaluation. For example, it may be the case that a logistic regression model performs similarly on both synthetic and real data, but that might not be the case when other models are used, such as SVMs or neural networks. To mitigate this issue, the authors of~\cite{jordon2018pate} propose Synthetic Ranking Agreement (SRA), a framework that evaluates a selection of models trained on the synthetic and tested on the synthetic. The performances of the same models are compared to those trained and tested on real data~\cite{jordon2018pate,jordon2020synthetic}. The authors then define a metric that performs ranking agreement and comparison to evaluate the power of the synthetic data for machine learning downstream tasks. Although this metric can suffer from the same limitation of TSTR and TRTS frameworks, it evaluates a broader range of machine learning classifiers which is a step closer to the ideal machine learning utility assessment.

\begin{align*}
\boldsymbol{SRA}\left(\left\{A_{i}\right\}_{i=1}^{L},\left\{C_{i}\right\}_{i=1}^{L}\right)=\frac{1}{L(L-1)} \sum_{j=1}^{L} \sum_{k \neq j}
\mathbb{I}\left(\left(A_{j}-A_{k}\right) \times\left(C_{j}-C_{k}\right)>0\right)
\end{align*}

Where $L$ is a set of predictive models $f_{1}, f_{2}, \ldots, f_{L}^{2}$.  $A_{i} \in \mathbb{R}$ stands for the performance of the models when trained and tested on the real data, while $C_{i} \in \mathbb{R}$ stands for the performance of the models trained and tested on the synthetic data~\cite{jordon2018pate}. To scale the valuation of the utility of synthetic data for machine learning applications,~\cite{yale2020generation} studied the educational utility by hosting an online challenge for students to evaluate the quality of the data. 

It is important to note, that some works applied one of the frameworks mentioned here, for example, TSTR, however, they did not explicitly mention the framework's name.  In many machine learning applications, synthetic data can be used to augment real data. To evaluate how much synthetic data is needed to achieve the desired performance, ~\cite{Scgan} presented a Data Augmentation Test, where the authors evaluated the synthetic data's utility for machine learning applications. Similarly, in the performance of models was evaluated using augmented dataset, while varying the percentages of synthetic data used in each variation~\cite{li2021generating}.

Data utility metrics and tests were also employed in non-machine learning tasks, to evaluate the synthetic data for its intended utilization. This was specifically seen in GANs for missing data imputation tasks where the GAN-imputed data was evaluated in terms of Root Mean Square Error (RMSE)~\cite{chai2014root}, Mean Absolute Error (MAE)~\cite{chai2014root} as shown in~\cite{camino2019improving,yoon2018gain,gupta2021concurrent}. GANs for imputation tasks were also evaluated post-imputation prediction performance in terms of AUROC, FI, and accuracy and benchmarked against other state-of-the-art data imputation techniques as seen in~\cite{yoon2018gain,yang2019categorical,zhang2018medical,gupta2021concurrent}. Similarly, GANs for estimating treatment effects work were evaluated in terms of Precision in Estimation of Heterogeneous Effect (PEHE), average treatment effect (ATE)~\cite{hill2011bayesian}, the average treatment effect on the treated (ATT)~\cite{shalit2017estimating}, and RMSE for controlling the confounding evaluation~\cite{li2020continuous}.
    

    
\subsection{\textbf{Qualitative Evaluation}}
Qualitative evaluation approaches are commonly utilized in GAN papers, to support the quantitative results with reasonable simplistic measures. For example, several papers reported visualization of data distributions and embeddings, such as comparing generated feature distribution plots~\cite{yale2020generation}, and correlation heat-maps~\cite{che2017boosting}. While others qualitatively compared patient trajectories by visually comparing the synthetic time-series signals~\cite{esteban2017real,beaulieu2019privacy,Scgan}. An example of a qualitative privacy evaluation is the interpolation test proposed by~\cite{esteban2017real}, where a pair of training are back-projected into the latent space and linearly interpolating them produces smooth variation in the sample space, then the GAN model is then used to produce samples at each point. The variation in the outputs is used as a proof of the GANs' ability to capture the distribution without memorizing the training samples~\cite{esteban2017real}. 

    Researchers in machine learning often conduct ablation studies, where different components of the model are removed to evaluate the effect of the ablated component on the synthetic data. 
    This kind of evaluation was also seen in~\cite{gupta2021concurrent} to understand the role of the time-series classification layer, and in~\cite{li2018semi} to measure the effect of semi-supervised learning branch on the performance. In \cite{li2021generating}, experiments for ablation studies are designed to evaluate the validity of network components for latent mapping and sequence generation.
    It is worth noting that several papers ablated various components in their model, without mentioning the ``ablation studies" as seen in~\cite{yan2020generating,yoon2018ganite,yan2020generating}.
    
Clinical validity and trust of the synthetic data is a major concern and a bottleneck in using synthetic data for clinical research. To address this, some papers conducted clinician evaluations, where a group of medical professionals are shown the data and asked to evaluate it based on its realism~\cite{lee2020generating,beaulieu2019privacy,choi2016medical,weldon2021generation}. The exact evaluation performed by clinicians can vary. For example, in~\cite{beaulieu2019privacy,choi2016medical}, the clinical evaluation team was asked to give a numerical rating (from 1 to 10) of the realism of the data. Other authors asked the clinical evaluation team to classify data encounters as either real or generated and used more qualitative rating scales such as ``Highly Plausible, Plausible, Implausible".~\cite{weldon2021generation}. The results of the clinical evaluations were then compared and reported using statistical metrics used for classification and statistical significance tasks.  To measure the GAN model's ability to obey clinical constraints among variables, Constraint Violation Test (CVT) was introduced where the authors calculated the differences between (max-median) and (median-min) for vital sign measurements in a tabular EHR setting~\cite{yan2020generating}. The difference values on the record level were calculated, where the signs and magnitude of the difference are indicative of the constraint violations~\cite{yan2020generating}. It is important to point out that the results from such qualitative techniques can be useful, but are not sufficient to provide conclusive measures on the performance and quality of the GAN-based models.

\section{\textbf{Open Access Data Sources}}
\label{sources}
To demonstrate their usefulness for EHR-related applications, the developed GAN-based models were trained on various EHR datasets, as shown in  Table \ref{tab:EHR_GAN}. The datasets vary in size, openness of access, included features, and recording settings. One of the most commonly used datasets for GANs for EHRs development is Medical Information Mart for Intensive Care \textit{(MIMIC) III}  \cite{johnson2016mimic}, which was collected from critical-care settings from Beth Israel Deaconess Medical Center (BIDMC) in the United States~\cite{johnson2016mimic}. Some of the included features were categorical and discrete such as demographics and patient outcomes. Others were continuous time-stamped vital-signs measurements, as well as clinical and imaging notes, and  interventions. Its free access,  extensive documentation, and online support community make it a suitable candidate for tabular and time-series GANs for EHRs applications. Another freely available data is \textit{Philips eICU}~\cite{pollard2018eicu}, which is a multi-center database for critical care data. Philips eICU was collected from more than 208 hospitals throughout the United States between 2014-2015, making it a good choice for validating models across multi-centers. Both MIMIC and eICU datasets can be freely downloaded from \textit{PhysioNet}, a resource that provides access to extensive collections of physiological and clinical data and related open-source software.~\cite{moody2001physionet}. Most \textit{PhysioNet} datasets are accessible to the public users following the registration and signing of a data use agreement, with some datasets requiring additional credentialing. A recently introduced ICU dataset that was also made available on \textit{PhysioNet} is \textit{HiRID}, a high time-resolution ICU dataset collected from an ICU in Switzerland~\cite{hyland2020early}. Similarly, another European ICU data is the Amsterdam University Medical Centers Database (AmsterdamUMCdb), which was released in 2021~\cite{thoral2021sharing}. HiRID and AmsterdamUMCdb are suitable datasets for critical care research for works interested in validating their models on populations outside the United States.  

Another openly available data source is the University of California Irvine (UCI) Machine Learning Repository, which maintains 588 data sets that can be used for a wide range of applications since 2007~\cite{asuncion2007uci}. The repository now includes several small medical datasets. Some examples of the UCI medical datasets include UCI Epileptic Seizure Recognition, UCI breast Cancer, and UCI Heart Disease datasets~\cite{asuncion2007uci}. When using UCI datasets for benchmarking, one should be mindful of the datasets' similar names. For example, six distinct datasets include the word “breast” in their names, each with a different number of features and type of variables~\cite{asuncion2007uci}. There's also a lack of standardization in the documentation of each of the datasets, since some have the patient identifiers, and target variables included as features, while others do not.  Careful and detailed documentation and reporting of the used dataset are essential to allow for accurate benchmarking and reproducibility. The development of data-science competitions such as the ones hosted on Kaggle and \textit{PhysioNet}, resulted in open access healthcare datasets that were used in GAN-based works such as the Kaggle Cervical Cancer and The PhysioNet Challenge 2012, as seen in~\cite{jordon2018pate} and~\cite{luo2018multivariate,luo2019e2gan}, respectively.

A number of the reviewed works used RCT data, some of which are not directly accessible upon request and signing a user agreement. Notably, RCT datasets that have been used for several clinical research publications include Systolic Blood Pressure Intervention Trial (SPRINT)~\cite{sprint2015randomized}, and Meta‐Analysis Global Group in Chronic (MAGGIC)~\cite{wong2014heart} which includes data from 30 RCTs for patients with heart failure. When evaluating GANs for treatment effects, the benchmarking datasets used were the ones commonly used for causal inference applications in general. Notably, \textit{TWINs}~\cite{almond2005costs} dataset collected for births from 1989-1991 in the United States was used for binary treatment research, where twins data mimic the factual and counterfactual observations, for a certain outcome such as mortality within the first year of birth. Several covariates are recorded such as race, pregnancy period, and quality of care during pregnancy.  Another commonly used dataset for treatment effects is the Infant Health and Development Program (IHDP) data, first introduced by~\cite{hill2011bayesian}, which belongs to an RCT that began in 1985 focusing on premature infants and the efficacy of educational and family support services on the infants over a 3-year period of their life~\cite{brooks1992effects}. 

Other data sources such as Surveillance, Epidemiology and End Results (SEER) of the of the National Cancer Institute~\cite{jemal2002cancer}, Nemours Pediatrics longitudinal pediatric encounter-base data~\cite{gupta2019obesity}, and the United Network for Organ Transplantation (UNOS)
~\cite{cecka1992unos}, can be obtained upon request from their dedicated websites.  There are several other referenced datasets in the literature, however, those were private and not accessible for open access GANs for EHRs development.  

\section{\textbf{Future Outlook}}
\label{next}

The recent developments of GANs for EHRs are promising first steps for potential research and decision support systems applications. The works we have reviewed in this paper reveal many opportunities for developments in theory, algorithms, and applications.
However, we believe that there are some challenges and gaps that need to be addressed and taken into consideration. 


\subsection{\textbf{Evaluation of Synthetic Data}}

The lack of a universal evaluation methodology is a bottleneck in developing reliable GANs for EHRs works. As shown in Table~\ref{tab:EHR_GAN}, there is no standardization in the evaluation components or the metrics. Currently, researchers tend to either (1) use commonly used metrics for GAN applications in other fields such as imaging and non-medical time-series, (2) use metrics utilized by benchmark models, or (3) introduce their own new metrics. In addition, we noticed that the same evaluation test is referred to using different names in some cases, which adds to the confusion regarding GAN evaluation~\cite{lee2020generating,zhang2021synteg}. 
When evaluating the machine learning utility, we believe that it is essential to report the results on both the synthetic and real datasets to understand the model's baseline performance and accurately determine the utility of the synthetic data for downstream tasks. We note that different metrics can lead to various limitations and trade-offs. Therefore, currently, it is hard to determine the state-of-the-art GANs for EHRs models.   While we believe that providing qualitative evaluations and analysis adds value to the studies, it is insufficient without supporting rigorous quantitative evaluations. In this work, we categorized the metrics based on the data aspect they are evaluating and whether they can be applied to each type of EHRs data. We hope that our work inspires future investigations of the newly introduced evaluation tests' strengths, limitations, and trade-offs to standardize a guideline for selecting the metrics and their weights and matching them to the synthetic data utility. 

Furthermore, we believe that future research should investigate aspects related to the general utility of the data should also be considered in the optimization criteria. For instance, synthetic data generated for data augmentation in machine learning tasks should be evaluated differently than generating research purposes or imputing missing values and estimating counterfactuals, which might go beyond the predictive utility of the data. In the current literature on GANs for EHRs, there is no clear path to how the generated data is disseminated beyond the scope of research hypothesis testing setups. To this end, GAN training is computationally expensive and can lack stability; therefore, we recommend that future works evaluate computational complexity to allow for lightweight GAN development and dissemination. 

\subsection{\textbf{Privacy-Similarity Trade-off}}
The principles of GANs' architecture rely on the competing goals of the generator and discriminator, which overall optimize for the synthetic distribution similarity. The synthetic nature of GANs outputs implicitly preserves privacy since there is no direct mapping between a single synthetic output and a real input. However, unintentional information leakage can ensue when dealing with sensitive information such as EHRs, as shown in the previously discussed adversarial attack mechanisms. The privacy-similarity trade-off was a recurring theme in various works. We believe that to address the similarity-privacy trade-off dilemma, authors should test for both factors irrespective of the chosen level of privacy guarantees. We observe that some of the early works did not consider testing for information leakage risks.
Similarly, some of the works focusing on privacy improving privacy-preservation of the GAN models did not adequately evaluate the data for preserving the distribution similarity. Conservative privacy guarantees such as differential privacy are helpful but can have a high cost on the fidelity and utility aspects. Considering that such strict differential privacy guarantees are not required by GDPR nor HIPPA for medical applications, we advise for at least considering one of the more relaxed privacy preservation evaluation techniques.  We believe that future directions of research should work with regulatory bodies to establish a clear guideline on the privacy risks to allow for private data owners to share synthetic data with confidence, which will open the doors for a wave of new research applications.

\subsection{\textbf{Generation of EHRs from Multimodal Data and Multi-Centers}}

The diversity resulting from the collection of various clinical information opens the doors for various research data-driven models. For example, as shown in section~\ref{data_gen}, various GAN models were developed to generate different EHR data types such as tabular snapshot during patient's encounter (such as diagnosis and procedure ICD codes), as well as clinical time-series collected over time (such as vital signs and laboratory measurements). However, very few works investigated generating data that captures the correlations between heterogeneous types of data, i.e., simultaneously generating EHR data with different types while modeling their underlying relationships \cite{li2021generating}. 
Furthermore, even though we limited the scope of this review to structured EHRs, in real-world applications, medical data comes in other modalities such as unstructured clinical notes and medical imaging documented in EHR systems, which have related areas in natural language processing (NLP) \cite{ruppel2020assessment} and computer vision \cite{esteva2021deep} research. 
Leveraging information existing in EHRs with mixed modalities can help GANs generate patient records with higher fidelity. Moreover, generating EHR data from a holistic perspective can also contribute to the realization of the concept of `digital twins' and personalised medicine in the future \cite{georges2020synthetic,georges2020synthetic}.

Training deep neural networks requires large amounts of data, that is representative of the target patient population, which usually entails training on data from multiple institutions. Despite using multiple datasets, the majority of the papers reviewed in this work train on one dataset at a time. We believe that researchers first must overcome the challenges of feature mismatch and distribution mismatch~\cite{yoon2018radialgan}, to reach the optimal application of a GAN model in different institutions. The literature is still nascent with respect to applications of GANs for EHRs for implementation in different institutions. One of the few works that explored the use of GANs for domain translation to facilitate the use of EHR data from multi-centers, was \textit{RadialGAN}~\cite{yoon2018radialgan}. Recently, the first GAN-based federated learning framework for tabular EHRs was proposed~\cite{weldon2021generation}. However, the authors only used a single dataset and split it into separate data silos in an experiment to simulate multi-centers. Separate GAN models were trained on each of the silos, and then were later combined in a central GAN model ~\cite{weldon2021generation}. We expect future works to investigate the feasibility and introduce new ways to implement GAN models on datasets from different institutions, and explore new applications such as federated and continual learning~\cite{armstrong2021continual} .

\subsection{\textbf{Reporting and Open Access Resources}}
Transparent reporting of training and validation datasets, preprocessing steps, hyper-parameter space, and training methodology in GAN-based applications are paramount for achieving safe use of the GAN model and for benchmarking and reproducibility of results. As noted by~\cite{camino2020working}, feature encoding techniques and entire hyper-parameter space are often not described despite having a substantial impact on the results for missing value imputation tasks. Without transparent and comprehensive reporting,  it becomes difficult to understand the GAN model's assumptions and limitations which then impedes their safe deployment and usage. 

On the other hand, we observe a positive trend of open access work, where most of the reviewed papers published their code online. Nevertheless, some papers mentioned providing open access code while lacking or referencing non-functional links. The open access datasets mentioned in section \ref{sources} allow for a wide range of GANs for EHRs applications. However, we also acknowledge that despite the usefulness of critical care and small-sized datsets a in many healthcare applications, their utility is limited in some tasks. For example, generating synthetic longitudinal data is important to study prescription activities, long-term treatments effects, and other population-wide research questions. Without open access datasets of different kinds, it will be challenging to expand GANs for EHRs research to include longitudinal data. 

\subsection{\textbf{Integration in Clinical Applications}}

Using simulated data in medical practice is not new; senior academic medics compile hand-engineered simulated data to train medical students and residents as a part of their education~\cite{cleland2009use}. However, using machine learning-based generated synthetic data for research and clinical support system raises concerns and questions of trust, reliability, and realism from the clinical research community. Currently, most quantitative evaluation tests and metrics are hard to interpret by medical professionals~\cite{chen2021synthetic}, which results in a gap between synthetic data and GAN development and their usage in clinical applications. To mitigate this gap, there is a need to develop evaluation tests that confirm the preservation of unique characteristics of clinical datasets that clinicians easily understand. 
We believe that using such metrics in conjunction with rigorous mathematical and statistical similarity evaluation will support the acceptance of the use of synthetic data. Furthermore, co-designing algorithms with clinicians 
generally enhances the field of machine learning to develop new architectures for various applications in healthcare.

With the increased number of works introducing new methodologies, evaluation metrics, and applications of GANs for EHRs, we believe that many of these models need to be validated on real-world large-scale EHR databases. By validating the included works on real-world EHR databases, we get a better understanding of the true scalability and reproducibility of data fidelity, utility and privacy results.  Furthermore, such validation is needed to test for the GANs' ability to capture variations of complex dependency relationships between variables stored in EHR databases from diverse clinical settings. 

We believe that synthetic data has the potential to inspire a wide range of clinical research as seen in non-GAN based synthetic datasets~\cite{CPRD_synthetic,CPRD_synthetic2,walonoski2020synthea}.  With reduced time for data access and ethics approvals, as seen in~\cite{guo2020use}, research can be expedited, supporting the advancement of machine learning for healthcare.
Overall, GANs for EHRs is a relatively new field and still has lots of capacity for improvement, especially in addressing EHR data complexity aspects such as heterogeneity, missingness, and sparsity. 

\section*{\textbf{Acknowledgments}}
JL was funded by the China Scholarship Council from the Ministry of Education of
P.R. China. TZ was supported by the RAEng Engineering for Development Research Fellowship and the National Institute for Health Research Oxford Biomedical Research Centre. 
\section*{\textbf{Author Contributions}}
GG and TZ conceived the concept of the paper. GG, JL, and TZ contributed to the data collection, literature review and discussions leading to the presented ideas. TZ provided supervision and guidance to the work. GG, JL, and TZ contributed to the writing and editing of the manuscript.
\section*{\textbf{Competing Interests}}
The Authors declare no Competing Financial or Non-Financial Interests.

%
\printbibliography

\end{document}